\documentclass[runningheads]{llncs}

\usepackage{eccv}
\usepackage{eccvabbrv}
\usepackage{graphicx}
\usepackage{booktabs}
\usepackage{bbding}      
\usepackage{colortbl}
\usepackage{multirow}
\usepackage{tabu}
\usepackage{pifont}
\usepackage{enumitem}
\usepackage{float}
\usepackage{setspace}
\usepackage[square,sort,comma,numbers]{natbib}
\usepackage{hyperref}
\usepackage{orcidlink}
\usepackage{graphicx}
\usepackage{booktabs}
\usepackage[accsupp]{axessibility} 
\definecolor{LightBlue}{rgb}{0.85, 0.92, 1.0} 

\makeatletter
\def\thanks#1{\protected@xdef\@thanks{\@thanks
        \protect\footnotetext{#1}}}
\makeatother

\begin{document}

\title{Cross-Domain Few-Shot Object Detection via Enhanced Open-Set Object Detector}

\titlerunning{CDFSOD with Enhanced Open-Set Object Detector}

\author{Yuqian Fu$^{*}$$^{1,2,3}$, Yu Wang$^{*}$$^{1}$, Yixuan Pan$^{*}$$^{4}$, Lian Huai$^{\dagger}$$^{5}$, Xingyu Qiu$^{1}$, Zeyu Shangguan$^{5}$,  Tong Liu$^{5}$, Yanwei Fu$^{1}$, Luc Van Gool$^{2,3}$, Xingqun Jiang$^{5}$
\thanks{*: equal contributions.}
\thanks{$\dagger$: corresponding author.}
\thanks{Part of this work commenced during Dr. Yuqian Fu's PhD at Fudan University.}
}
\authorrunning{Fu et al.}

\institute{
	$^1$Fudan University, China, $^2$ETH Zürich, Switzerland, $^3$INSAIT, Sofia University, "St. Kliment Ohridski”, Bulgaria, $^4$Southeast University, China, $^5$BOE Technology, China \\
	\email{yuqian.fu@insait.ai}, \email{yu\_w13@fudan.edu.cn}, \email{yixuanpan@seu.edu.cn}}

\maketitle

\begin{abstract}
This paper studies the challenging cross-domain few-shot object detection (CD-FSOD), aiming to develop an accurate object detector for novel domains with minimal labeled examples. While transformer-based open-set detectors, such as DE-ViT, show promise in traditional few-shot object detection, their generalization to CD-FSOD remains unclear: 1) can such open-set detection methods easily generalize to CD-FSOD? 2) If not, how can models be enhanced when facing huge domain gaps? To answer the first question, we employ measures including style, inter-class variance (ICV), and indefinable boundaries (IB) to understand the domain gap. Based on these measures, we establish a new benchmark named CD-FSOD to evaluate object detection methods, revealing that most of the current approaches fail to generalize across domains. Technically, we observe that the performance decline is associated with our proposed measures: style, ICV, and IB. Consequently, we propose several novel modules to address these issues. First, the learnable instance features align initial fixed instances with target categories, enhancing feature distinctiveness. Second, the instance reweighting module assigns higher importance to high-quality instances with slight IB. Third, the domain prompter encourages features resilient to different styles by synthesizing imaginary domains without altering semantic contents. These techniques collectively contribute to the development of the Cross-Domain Vision Transformer for CD-FSOD (CD-ViTO), significantly improving upon the base DE-ViT. Experimental results validate the efficacy of our model. Datasets and codes are available at \href{http://yuqianfu.com/CDFSOD-benchmark}{http://yuqianfu.com/CDFSOD-benchmark}.
\keywords{Cross-Domain Few-Shot Learning \and Few-Shot Object Detection \and Open-Set Detector}
\end{abstract}

\section{Introduction}
\label{sec:intro}

Cross-domain Few-Shot Learning (CD-FSL)~\cite{guo2020broader} 
deals with learning across different domains, transferring knowledge from a source dataset to new target domains with few labeled data. 
While many methods focus on the classification task in CD-FSL, they often overlook  \textit{object detection}. This paper delves into object detection tasks in CD-FSL, also known as Cross-Domain Few-Shot Object Detection (CD-FSOD).

\begin{figure}[t]
\centering	{\includegraphics[width=0.95\linewidth]{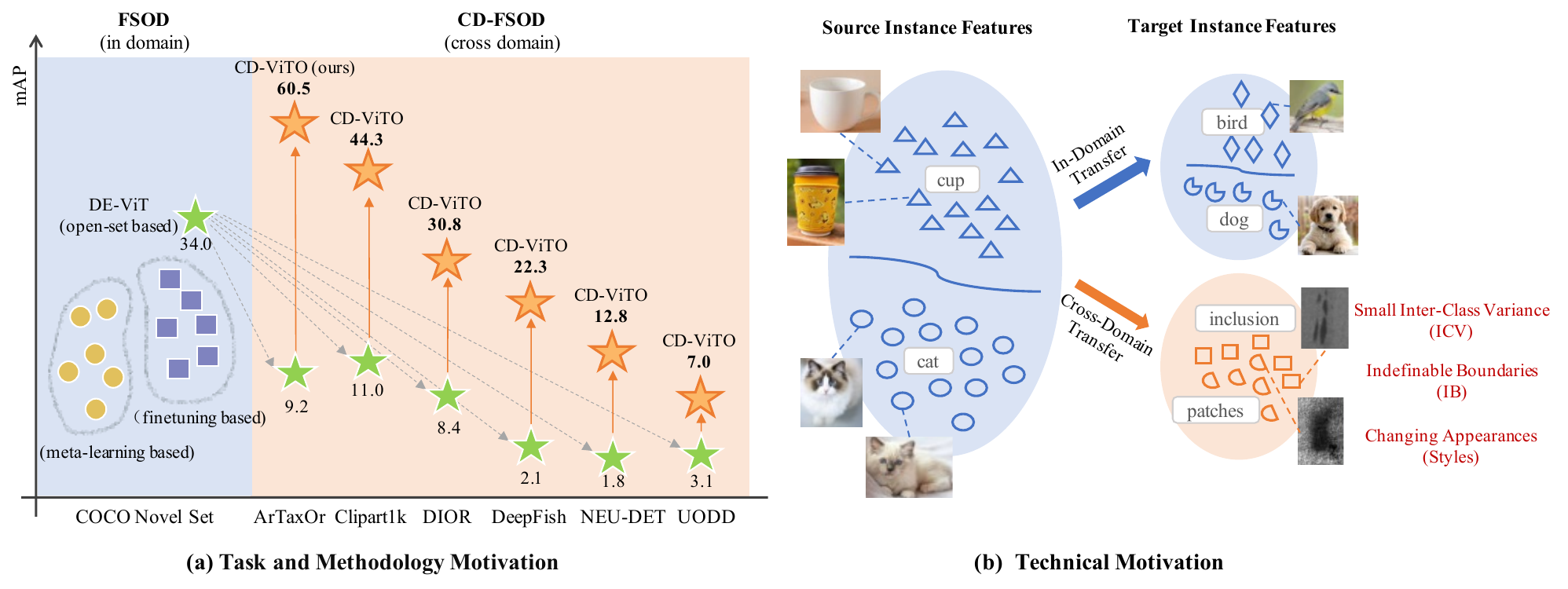}}
\caption{
 \small 
 (a) Our motivation: The DE-ViT open-set detector excels in FSOD but struggles in CD-FSOD, inspiring our creation of CD-ViTO. (b) Technical motivation: FSOD models face challenges when dealing with cross-domain targets, such as small inter-class variance (ICV), indefinable boundaries (IB), and varying appearances (styles). 
\label{fig:teaser} 
}
\vspace{-0.15in}
\end{figure}

CD-FSOD is derived from Few-Shot Object Detection (FSOD)~\cite{kohler2021few}, aiming to detect new objects with limited instances. Much effort has been dedicated to this objective. Current FSOD methods fall into two main categories: meta-learning based methods~\cite{yan2019meta, han2021query, han2022few}, which train a meta-learner adaptable to new objects, and finetuning based approaches~\cite{wang2020frustratingly, sun2021fsce, qiao2021defrcn, ma2023digeo, kaul2022label}, involving initial training followed by finetuning on few labeled data. Notably, DE-ViT~\cite{zhang2023detect}, an open-set object detector, shows exceptional performance in FSOD, surpassing other methods as depicted in Fig.~\ref{fig:teaser}(a).
This motivates us to explore if the detector is effective for the CD-FSOD task, posing two main questions: \ding{172} Can open-set detection methods generalize to cross-domain target datasets without performance degradation? \ding{173} If not, how can the open-set methods be improved in the presence of significant domain gaps?

To address question \ding{172}, we establish a new benchmark for CD-FSOD by rearranging existing object detection datasets. 
Our benchmark utilizes COCO~\cite{lin2014microsoft} as source training set and incorporates six additional datasets, namely ArTaxOr~\cite{GeirArTaxOr}, Clipart1K~\cite{inoue2018cross}, DIOR~\cite{li2020object}, DeepFish~\cite{saleh2020realistic}, NEU-DET~\cite{song2013noise}, and UODD~\cite{jiang2021underwater}, as novel target datasets.
We also introduce metrics such as style, Inter-Class Variance (ICV), and Indefinable Boundaries (IB) to measure domain differences.
Style represents different visual appearances, ICV measures dissimilarity between semantic labels, and IB reflects confusion between targets and backgrounds. 
Our target datasets exhibit significant variations in style, ICV, and IB, making our benchmark a comprehensive testbed for evaluating CD-FSOD methods. 
We evaluate various object detectors on this benchmark. As shown in Fig.~\ref{fig:teaser}(a), we observe obvious performance degradation even for the State-Of-The-Art (SOTA) FSOD detector, DE-ViT. This indicates the challenge of tackling CD-FSOD even with well-designed open-set detectors.

To tackle the performance degradation issue, i.e., question \ding{173}, 
we aim to upgrade the vanilla DE-ViT to improve the open-set detector's performance. 
Our technical motivation, illustrated in Fig.~\ref{fig:teaser}(b), highlights that FSOD model degradation correlates with domain metrics we mentioned earlier: style, ICV, and IB. Specifically, a model trained on instances with coarse semantic concepts (large ICV) and clear foreground-background boundaries (slight IB), like "cup" and "cat", can easily transfer to similar in-domain concepts like "bird" and "dog". However, cross-domain cases, such as industrial defect types like "inclusion" and "patches", pose challenges with smaller ICV, more significant IB, and different styles. This insight inspires us to design a method that addresses issues of small ICV, significant IB, and changing styles in cross-domain scenarios.

To achieve this, we propose several novel modules: learnable instance features, an instance reweighting module, and a domain prompter.
(1) Initially, to counter the negative effects of small ICV, we aim to increase feature discriminability making the target classes more distinguishable. This is achieved by our learnable instance features which adapt fixed initial instance features using target labels, aligning the features with semantic concepts. 
With explicit categories as supervision, the optimized features naturally exhibit greater dispersion than directly extracted ones. 
(2) Next, we introduce the instance reweighting module to assign different weights to instances, prioritizing those with slight IB. By emphasizing high-quality instances in forming class prototypes, we mitigate challenges posed by IB. 
(3) Further, to boost model robustness across domains, we propose a domain prompter to synthesize virtual "domains" and ensure prototype features remain stable under domain perturbations. 
To encourage the uniqueness of each synthesized domain, a domain diversity loss is proposed. Simultaneously, a contrastive learning loss and a classification loss are incorporated to maintain semantic content consistency in the presence of domain perturbations. 

Furthermore, we also explore the finetuning technique. Although finetuning has been studied in cross-domain tasks~\cite{guo2020broader, fu2023styleadv}, we provide a detailed analysis of its effectiveness on our benchmark.  
By integrating these novel modules into DE-ViT, we introduce a new vision transformer based open-set detector called CD-ViTO for CD-FSOD. 
CD-ViTO substantially boosts DE-ViT’s performance across all target datasets, as shown in Fig.~\ref{fig:teaser}(a) (highlighted by the orange stars).

We summarize our contributions. 
1) We propose a comprehensive benchmark for CD-FSOD that spans diverse target domains. Additionally, we introduce several metrics related to domain gap analysis, including style, inter-class variance, and indefinable boundaries.
2) Through extensive evaluations, we uncover the challenges posed by the domain gap issue in existing open-set object detectors and other object detection methods for CD-FSL.
3) Leveraging the existing DE-ViT, we present CD-ViTO, a novel method that incorporates finetuning, learnable instance features, an instance reweighting module, and a domain prompter. Notably, these new modules introduce minimal parameters and require only a few target support instances for training, ensuring high efficiency.
4) Our results demonstrate that CD-ViTO significantly improves upon the base DE-ViT, achieving new state-of-the-art performance in CD-FSOD.

\section{Related Works\label{sec:related}}
\noindent\textbf{Few-Shot Object Detection.}
FSOD~\cite{kohler2021few} identifies objects with few labeled data. Existing FSOD methods can be roughly divided into meta-learning based~\cite{yan2019meta, han2021query, han2022few, kang2019few, han2022meta} and transfer learning based~\cite{wang2020frustratingly, sun2021fsce, fan2021generalized, qiao2021defrcn, ma2023digeo, kaul2022label, guirguis2023niff}. 
Meta-learning trains model across tasks for quick adaptation. Meta-RCNN~\cite{yan2019meta} is a typical work of this type.  
Transfer learning finetunes pretrained model on limited data for few-shot tasks. Flagship works include TFA~\cite{wang2020frustratingly}, FSCE~\cite{sun2021fsce}, and DeFRCN~\cite{qiao2021defrcn}. 
Notably, a recently proposed method, DE-ViT~\cite{zhang2023detect}, unifying the problem of FSOD and open-vocabulary object detection~\cite{zareian2021open} and constructing an open-set detector for both tasks, shows great advantage. 
However, almost all of the methods mentioned above ideally assume that the training and testing sets belong to the same domain. 
Exceptions include Distill-cdfsod~\cite{xiong2023cd} and MoFSOD~\cite{lee2022rethinking}. Distill-cdfsod~\cite{xiong2023cd} introduces several datasets and a distillation-based method, but both the benchmark and method have room for improvement. MoFSOD~\cite{lee2022rethinking} also explores CD-FSOD and studies various model architectures, finetuning, and pre-trained datasets. 
Compared with them, apart from the datasets, we for the first time propose benchmark metrics and study the open-set detectors for CD-FSOD. Note that AcroFOD~\cite{gao2022acrofod} and AsyFOD~\cite{gao2023asyfod} also explore FSOD across domains. However, they share the same class set between source and target domains, while we handle entirely novel target categories.

\noindent\textbf{Cross-Domain Few-Shot Learning}.
CD-FSL aims to improve few-shot learning (FSL)~\cite{snell2017prototypical, vinyals2016matching, zhang2023deta, luo2023closer, TangYLT22}  across diverse domains. While there are several existing works~\cite{tseng2020cross, guo2020broader, wang2021cross, fu2021meta, zhuo2022tgdm, hu2022pushing, fu2023styleadv} addressing classification tasks, there is a noticeable gap in tackling domain gap challenges across various few-shot vision tasks, including object detection~\cite{yan2019meta, han2021query}, semantic segmentation~\cite{wang2019panet, xie2021few}, and action recognition~\cite{zhang2020few, fu2020depth}. 
This paper specifically explores the area of CD-FSOD, seeking to develop a robust object detector.
Despite recent advancements~\cite{gao2022acrofod, xiong2023cd, lee2022rethinking} in CD-FSOD, certain crucial aspects, such as domain gap metrics, comprehensive baseline evaluations, and technical improvements, remain less explored. 
Thus, we meticulously organize the benchmark, conduct an in-depth study on various existing methods, and introduce CD-ViTO as a novel method. Our goal is to propel the field of CD-FSOD forward, shedding light on unexplored aspects.

 \noindent\textbf{Open-Vocabulary Object Detection (OVD).} 
 Traditional object detection models like Fast R-CNN~\cite{girshick2015fast}, YOLO~\cite{redmon2016you}, and ViTDet~\cite{li2022exploring} recognize a predefined set of object categories. OVD~\cite{zareian2021open} aims to extend this recognition to novel categories by leveraging the shared space between images and category text. 
 This shared space can be established by the method itself~\cite{zareian2021open} or by utilizing existing vision-language frameworks like CLIP~\cite{radford2021learning}, as shown in RegionCLIP~\cite{zhong2022regionclip}, VL-PLM~\cite{zhao2022exploiting}, and Detic~\cite{zhou2022detecting}.
 Particularly, DE-ViT~\cite{zhang2023detect} is an open-set detector\footnote{\small 
Despite varying definitions, we adopt the one of~\cite{zhang2023detect} that  ``both open-vocabulary and few-shot belong to open-set except their category representations''.} capable of detecting objects of arbitrary classes.  Despite DE-ViT relying solely on the visual modality without using the text modality in practice, its motivation, method design, and experimental validations align with the OVD scope. Here, we made an exploration of open-set object detectors for CD-FSOD.

\section{CD-FSOD: Setup, Metrics, and Benchmark}
\subsection{Task Configurations }
\label{sec:trainingandinfer}

\noindent \textbf{Task Formulation. }\label{sec:formulation}
Formally, given a source dataset denoted as $\mathcal{D}_{S} = \{I, y\}, y \in \mathcal{C}_{S}$ with distribution $\mathcal{P}_{S}$ and a novel target dataset $\mathcal{D}_{T} = \{I, y\}, y \in \mathcal{C}_{T}$ with distribution $\mathcal{P}_{T}$, consistent with FSOD, CD-FSOD supposes the labeled data for the source classes $\mathcal{C}_{S}$ is sufficient while each novel target class in $\mathcal{C}_{T}$ has only a few labeled instances. All target classes must be novel to model i.e., $\mathcal{C}_{S} \cap \mathcal{C}_{T} = \emptyset$. 
In addition, different from the FSOD which assumes the distribution of source $\mathcal{P}_{S}$ equals that of the target $\mathcal{P}_{T}$, CD-FSOD tackles a more realistic scenario where $\mathcal{P}_{S} \neq \mathcal{P}_{T}$.  Models are required to train on $\mathcal{D}_{S}$ and then test on $\mathcal{D}_{T}$.

\noindent \textbf{$N$-way $K$-shot Protocol.} To evaluate the FSL ability of CD-FSOD models, we follow the $N$-way $K$-shot evaluation protocol. Specifically, for each novel class in $\mathcal{C}_{T}$, $K$ labeled instances are provided which is also termed as support set $S$, and other unlabeled instances are used as query set $Q$. Formally, we have $\lvert S \rvert = N \times K, N = \lvert \mathcal{C}_{T} \rvert$. 

\noindent \textbf{Training Strategy.} 
We adopt a "pretrain, finetune, and testing" pipeline, as used in many previous CD-FSL methods~\cite{guo2020broader, hu2022pushing}. 
This pipeline involves training methods on the source dataset $\mathcal{D}_{S}$, followed by finetuning the trainable parameters using the few labeled support instances $S$ from target dataset $\mathcal{D}_{T}$, and finally testing on the novel query set $Q$. 
Typical objective functions for training and finetuning include the box regression loss and the classification loss.

\subsection{Metrics for Domain Difference}

We aim to comprehensively capture challenges in cross-domain scenarios by evaluating the datasets across the following three dimensions.

\noindent \textbf{Style}:
We recognize style's pivotal role in various domain-related tasks like domain adaptation~\cite{luo2020adversarial}, domain generalization~\cite{zhou2021domain}, and cross-domain few-shot learning~\cite{fu2023styleadv}. Common styles include photorealistic, cartoon, 
sketch, etc.

\noindent  \textbf{Inter-class variance (ICV)}:  
ICV, a widely used metric in learning, measures the dissimilarity between classes. Higher ICV values indicate easier recognition of semantic labels. Coarser datasets like COCO usually have higher ICV, whereas finer-grained datasets exhibit smaller ICV values.

\noindent \textbf{Indefinable Boundaries (IB)}: 
IB, borrowed from camouflaged object detection~\cite{fan2021concealed}, reflects the confusion between the target object and its background. Greater confusion poses a challenge for object detectors. For example, detecting a person against a clean background is relatively straightforward, but identifying a fish in a coral reef is much more challenging.

We categorize ICV levels as large, medium, and small, and IB levels as slight, moderate, and significant. Specifics regarding the measurement of ICV and IB levels for datasets are provided in the supplementary materials.

\begin{figure*}
\centering	{\includegraphics[width=0.8\linewidth]{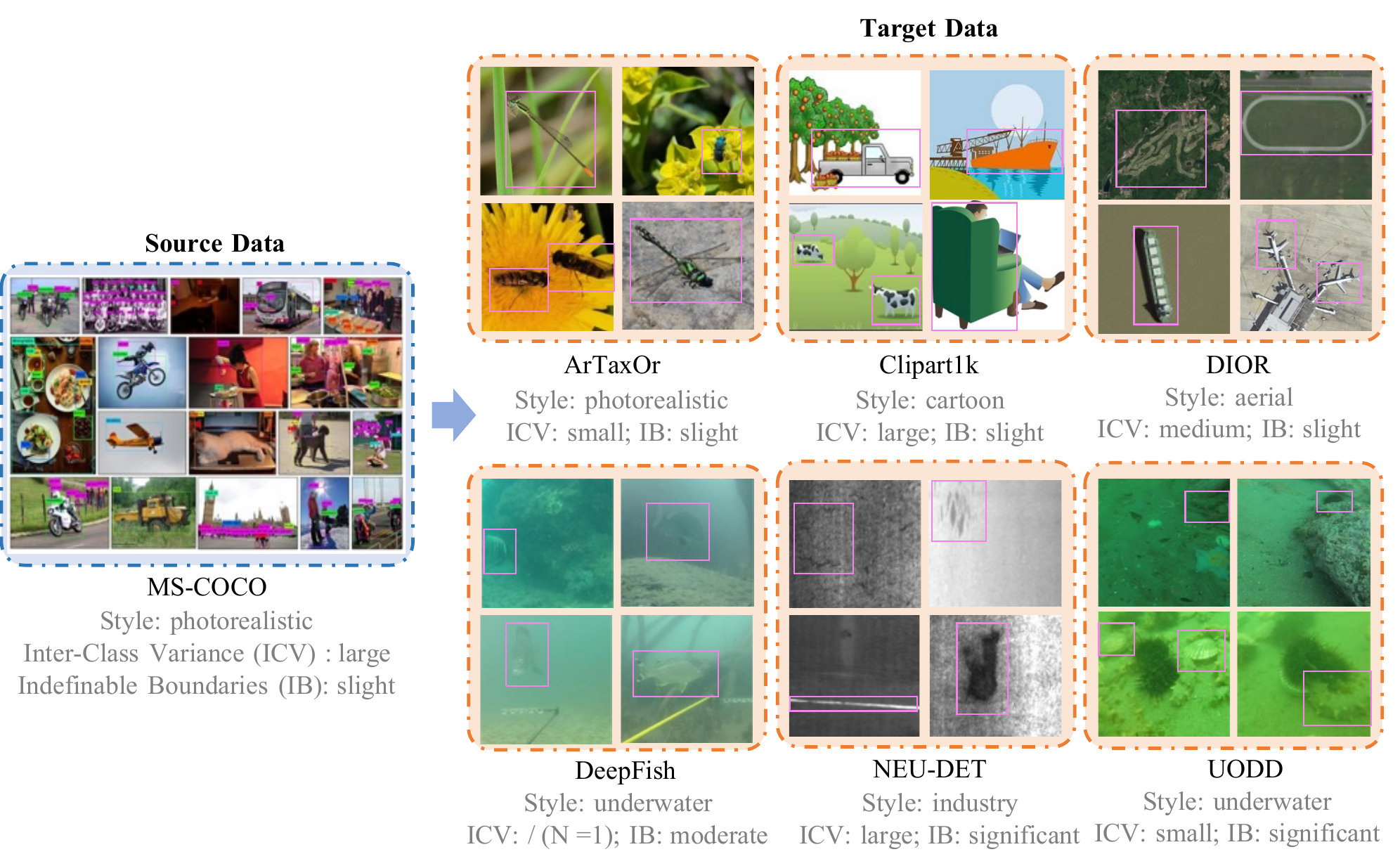}}
\caption{\small CD-FSOD benchmark: 
COCO serves as the training source data, while six datasets are utilized as novel testing target datasets.
These datasets exhibit variations in styles, inter-class variance (ICV), and indefinable boundaries (IB).
}\label{fig:benchmark} 
\end{figure*}

\subsection{Benchmarks of Different Domains}\label{sec:benchmark}

Based on the style, ICV, and IB metrics, we carefully review and reorganize existing object detection datasets to construct the CD-FSOD benchmark.
Our benchmark comprises seven datasets.
COCO~\cite{lin2014microsoft}, the widely used dataset for object detection, which provides a diverse set of object categories such as people, animals, vehicles, and various everyday objects serve as the $\mathcal{D}_{S}$. 
The other six datasets including ArTaxOr~\cite{GeirArTaxOr}, Clipart1k~\cite{inoue2018cross}, DIOR~\cite{li2020object}, DeepFish~\cite{saleh2020realistic}, NEU-DET~\cite{song2013noise}, and UODD~\cite{jiang2021underwater} function as $\mathcal{D}_{T}$. Examples of these datasets as well as the corresponding style, ICV, and IB are depicted in Fig.~\ref{fig:benchmark}. Note that DeepFish has only one class and thus hasn't an ICV value. More details of the datasets are attached to the supplementary materials.

\section{Methodology}
\label{sec:method}

\subsection{Overview of CD-ViTO} 

\begin{figure*}[t]
\centering	{\includegraphics[width=.8\linewidth]{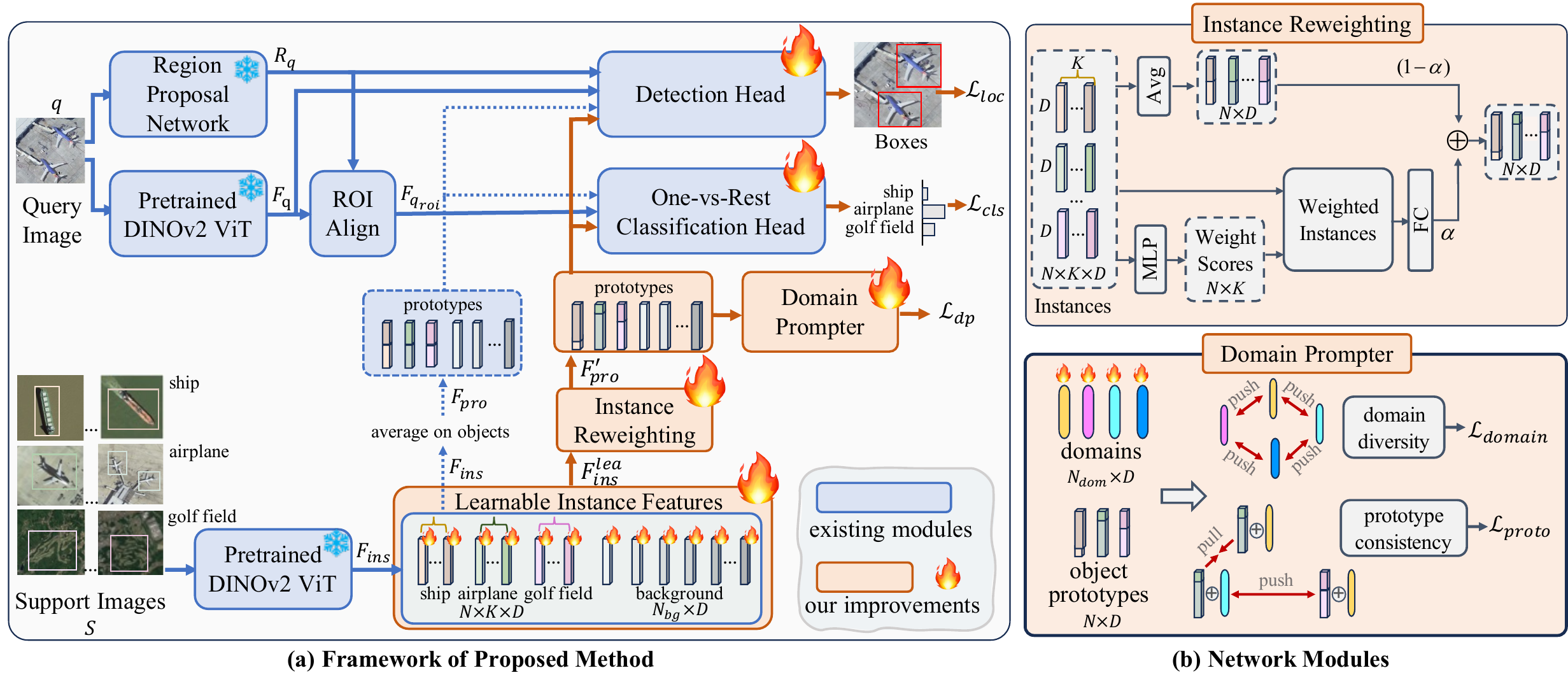}}
\caption{(a) Overall framework of our CD-ViTO. We build our method upon an open-set detector (DE-ViT). Modules in blue are inherited from DE-ViT while modules in orange are proposed by us. 
New improvements include learnable instance features, instance reweighting, domain prompter, and finetuning; (b) Illustration of our modules.\label{fig:framework} 
}
\end{figure*}

\noindent\textbf{Preliminary}. 
DE-ViT ~\cite{zhang2023detect} builds an open-set detector by using the visual features from large pretrained model and disentangling the localization and classification tasks.
The pipeline for the base DE-ViT is represented by blue arrows in Fig.~\ref{fig:framework}(a).
It mainly contains a pretrained DINOv2 ViT~\cite{oquab2023dinov2}, a region proposal network ($M_{RPN}$), an ROI align module ($M_{ROI}$), a detection head ($M_{DET}$), and a one-vs-rest classification head ($M_{CLS}$).
Particularly, given a query image $q$ and a set of support instances $S$, 
DE-ViT first uses the DINOv2 to extract the instance features $F_{ins} = \{{F_{ins}^{ob}, F_{ins}^{bg}}\}$, where the $F_{ins}^{ob}$ represents features for foreground objects from $S$ and the $F_{ins}^{bg}$ denotes features for background instances. 
Then, the class prototypes $F_{pro}= \{{F_{pro}^{ob}, F_{ins}^{bg}}\}$ are obtained by averaging $F_{ins}^{ob}$  on object categories while keeping $F_{ins}^{bg}$ unchanged (as in blue dashed arrows). 
For the query image $q$, 
DE-ViT applies the DINOv2, $M_{RPN}$, and $M_{ROI}$ to generate the region proposals $R_{q}$, visual features $F_q$, and ROI features $F_{q_{roi}}$. After that, the $M_{DET}$ takes the $R_{q}$, $F_q$, and $F_{pro}$ as input for the localization task producing $\mathcal{L}_{loc}$. In the meantime, the $M_{CLS}$ performs the classification task based on $F_{q_{roi}}$ and $F_{pro}$ resulting in $\mathcal{L}_{cls}$. The network is optimized by $\mathcal{L}_{loc}$ and  $\mathcal{L}_{cls}$.

Despite DE-ViT's purported ability to detect any novel objects without finetuning, our experiments reveal its failure to generalize to cross-domain target datasets (Sec.~\ref{sec:exp-cdfsod}). This motivates us to improve the cross-domain ability of DE-ViT and form a new method. Note that the basic components of DE-ViT including the DINOv2, $M_{RPN}$, $M_{ROI}$, $M_{DET}$, $M_{CLS}$, $\mathcal{L}_{loc}$, and $\mathcal{L}_{cls}$, validated on FSOD tasks, are directly inherited in our framework.
Our method, CD-ViTO, is outlined in Fig.~\ref{fig:framework}(a) with new enhancements highlighted in orange.
Recall that in Sec.~\ref{sec:intro}, we point out that the domain gap of CD-FSOD is highly related to the ICV, IB, and style of the target datasets. 
Therefore, we are inspired to design our CD-ViTO by tackling these factors. As a result, the novel \textit{learnable instance features} ($M_{LIF}$), the \textit{instance reweighting} ($M_{IR}$), the \textit{domain prompter} ($M_{DP}$) are proposed to tackle the small ICV, significant IB, and changing styles, respectively. In addition, though being explored in related cross-domain tasks~\cite{guo2020broader, fu2023styleadv}, the finetuning (FT) is also explored in our benchmark.

Specifically, given $S$ and $q$ as input, we perform the same steps as DE-ViT to obtain instance features $F_{ins} =\{F_{ins}^{ob}, F_{ins}^{bg}\}$, region proposals $R_{q}$, visual features $F_{q}$, and ROI features $F_{q_{roi}}$.
But different from directly averaging $F_{ins}^{ob}$ to obtain class prototypes $F_{pro}^{ob}$ for objects, we first utilize our $M_{LIF}$ to make the precalculated $F_{ins}$ learnable and adapt them as $F_{ins}^{lea}$ by performing the disentangled detection tasks i.e., the localization task supervised by $\mathcal{L}_{loc}$ and the classification task supervised by $\mathcal{L}_{cls}$. 
Since the training data is labeled with class categories, such an optimization would align the instance features to their corresponding categories thus increasing the discriminability of features and addressing the small ICV issue. 
Then, considering the qualities especially the IB of object instances varies, 
we apply our $M_{IR}$ to reweight the weights of objects in the $F_{ins}^{lea}$, this step forms the new prototypes $F_{pro}^{'}$.
Generally, $M_{IR}$ is expected to assign higher values to the high-quality instances with slight IB. In this way, the high-quality instances could contribute more to its class prototype. The $M_{IR}$ is inserted after the $M_{LIF}$ and is optimized use the same $\mathcal{L}_{loc}$, $\mathcal{L}_{cls}$.
In addition, to make the model more robust to the changing styles, our proposed $M_{DP}$ is applied to synthesize several virtual "domains" $F_{domain}$. The goal of the $M_{DP}$ is to introduce such domain perturbations and force our features to be consistent with different domains. To ensure the diversity of the introduced domains and constrain the domains would not affect the semantics of features, an extra loss $\mathcal{L}_{dp}$ is introduced. 
The $M_{DP}$ is build upon the prototypes reweighed by $M_{IR}$ i.e., $F_{pro}^{'}$,
and the $\mathcal{L}_{dp}$ works with the $\mathcal{L}_{loc}$ and $\mathcal{L}_{cls}$ together.

As for finetuning, we finetune the top two modules i.e., the $M_{DET}$ and $M_{CLS}$. We highlight that despite its simplicity, it is still worth exploring in the context of using open-set detectors for CD-FSOD. Experimental results (Sec.~\ref{sec:exp-cdfsod}) also demonstrate the necessity of our approach.
As a whole, our CD-ViTO tackles the cross-domain issues by setting instance features learnable, reweighting instances, adding domain prompter as useful perturbations, and finetuning key components. All the introduced parameters (indicated by the fire icons in Fig.~\ref{fig:framework}(a)) are very lightweight and we optimize them via the object function $\mathcal{L}$ as:
\begin{equation}\label{eq:loss}
\mathcal{L}  = \mathcal{L}_{loc} + \mathcal{L}_{cls} + \mathcal{L}_{dp}.
\end{equation}

\subsection{Key Components of Our CD-ViTO}

\noindent \textbf{Learnable Instance Features $M_{LIF}$.}
The implementation of our learnable instance features $M_{LIF}$ is quiet easy and efficient. 
Given the precalculated instance features $F_{ins} \in \mathcal{R}^{(N \times K + N_{bg}) \times D}$, where $N$, $K$, $N_{bg}$, $D$ represents the number of object classes, the number of shots, the number of background instances, and the dimension of visual features, $M_{LIF}$ defines a new learnable matrix $F_{ins}^{lea}$ with the same size of $F_{ins}$ and use the $F_{ins}$ for initialization. Formally, we have $F_{ins}^{lea} \in \mathcal{R}^{(N \times K + N_{bg}) \times D}, F_{ins}^{lea} = F_{ins}$. By optimizing the values of the $F_{ins}^{lea}$, we obtain the instance features better aligned to the target semantics. 
The idea of learnable instance features aligns with the visual prompt learning~\cite{jia2022visual} where the aim is to modify the input to better suit downstream tasks. In our case, we propose to modify the key instance features by setting them as learnable parameters. Most importantly, our module serves as the solution to making features more discriminative thus better tackling target data with small ICV.

\noindent \textbf{Instance Reweighting Module $M_{IR}$.}
We propose $M_{IR}$ to enable high-quality object instances e.g., objects with slight IB contribute more to the class prototype.
The architecture of $M_{IR}$ is shown in Fig.~\ref{fig:framework}(b). 
Given the learnable instance features $F_{ins}^{lea}$ from $M_{LIF}$, we first obtain the object instance features from  $F_{ins}^{lea}$ and denote it as $F_{ins}^{0} \in \mathcal{R}^{N \times K \times D}$. 
Our $M_{IR}$ has two paths connected in a residual way. 
More specifically, the lower path feeds $F_{ins}^{0}$ into an MLP module to get the weighted score which is further used to perform the weighted sum on the initial $F_{ins}^{0}$ resulting in the deformed prototypes $F_{pro}^{att} \in \mathcal{R}^{N \times D}$. The upper path performs the average operation on $F_{ins}^{0}$ getting $F_{pro}^{avg} \in \mathcal{R}^{N \times D}$. The final object prototypes $F_{pro}^{ob'}$ are represented as $\alpha fc(F_{pro}^{att}) + (1-\alpha)F_{pro}^{avg}$, where $fc(\cdot)$ is a fully connected layer and $\alpha$ is a hyperparameter. 
With $F_{pro}^{ob'}$ and background features, we form new prototype $F_{pro}^{'}$.

\noindent \textbf{Domain Prompter $M_{DP}$.}
To make our object prototype features $F_{pro}^{ob'}$ robust to various domains, we have to solve two goals: 1) synthesize several "domains" with diversities; 
2) prevent semantic drift when perturbing features with such "domains".
To achieve these, our $M_{DP}$ first introduces several learnable vectors $F_{domain} \in \mathcal{R}^{N_{dom} \times D}$ as virtual domains and then proposes several loss functions as supervision. 
The concept of "prompter" is also related but differs from the prompt tuning~\cite{jia2022visual, lester2021power}. 
Notably, both approaches employ trainable parameters as prompters, but general prompt tuning facilitates the transfer of pretrained models through efficient tuning, whereas we introduce prompters as perturbations.
The illustration of $M_{DP}$ is demonstrated in Fig.~\ref{fig:framework}(b). $N_{dom}$ is a hyperparameter. 

Concretely, given $F_{domain}$ and $F_{pro}^{ob'}$, we first propose the domain diversity loss $\mathcal{L}_{domain}$ to force the different domains e.g., $f^{d_{i}}$ and $f^{d_j}$ to be far from each other. Secondly, for a prototype $f_{p_{i}}$ in $F_{pro}^{ob'}$, we randomly perturbs it with two various domains $f^{d_{k}}$ and $f^{d_{m}}$ sampled from $F_{domain}$ forming two perturbed prototypes $f_{p_{i}}^{d_{k}}, f_{p_{i}}^{d_{m}}$. The perturbation is done by adding the features e.g., $f_{p_{i}}^{d_{k}} = f_{p_{i}} +f^{d_{k}}$. 
The domain $f^{d_{m}}$ is also added to other object prototypes forming $f_{p_{j}}^{d_{m}}$, where $j = 1, \ldots, N$. 
We constrain that $f_{p_{i}}^{d_{k}}$ and $f_{p_{i}}^{d_{m}}$ should be close to each other while being distant from perturbed prototypes generated from different class prototypes.
This constitutes our prototype consistency loss $\mathcal{L}_{proto}$.  Both the $\mathcal{L}_{domain}$ and $\mathcal{L}_{proto}$ are implemented by the InfoNCE loss as follows:
\begin{equation} 
\mathcal{L}_{domain}=-\frac{1}{N_{dom}}\sum_{i=1}^{N_{dom}}\left(\log \frac{\exp \left({f}^{{d}_{i}} \cdot {f}^{d_{i}} / \tau \right)}{\sum_{j=1}^{N_{dom}} \exp \left({f}^{{d}_{i}} \cdot {f}^{d_{j}} / \tau \right)}\right),
\end{equation}
\begin{equation}
\mathcal{L}_{proto}=-\frac{1}{N}\sum_{i=1}^{N}\left(\log \frac{\exp \left({f}_{{p}_{i}}^{d_k} \cdot {f}_{p_{i}}^{d_m} / \tau \right)}{\sum_{j=1}^{N} \exp \left({f}_{{p}_{i}}^{d_k} \cdot {f}_{p_{j}}^{d_m} / \tau \right)}\right), 
\end{equation}
where the $\tau$ is a temperature hyperparameter. Besides, we also classify the perturbed features e.g., $f_{p_{i}}^{d_{k}}$ into the corresponding class labels producing a new cross-entropy loss $\mathcal{L}_{{proto}_{cls}}$.  
Based on $\mathcal{L}_{domain}$, $\mathcal{L}_{proto}$, and $\mathcal{L}_{{proto}_{cls}}$, we have:
\begin{equation}
\mathcal{L}_{dp} =  \mathcal{L}_{domain} +  \mathcal{L}_{proto} + \mathcal{L}_{{proto}_{cls}}.
\end{equation}

\section{Experiments}\label{sec:exp}
\noindent\textbf{Datasets.} 
We perform CD-FSOD experiments using the benchmark introduced in Sec.~\ref{sec:benchmark}. In summary, the model is trained on COCO and evaluated on six other datasets. For ArTaxOr, DIOR, and UODD, we adhere to the splits used in Distill-cdfsod~\cite{xiong2023cd}. 
For Clipart1k, DeepFish, and NEU-DET, we create new splits for CD-FSOD evaluation.

\noindent\textbf{Network Modules.} 
All the modules inherited from the base DE-ViT including the pretrained DINOv2 ViT, $M_{RPN}$, $M_{ROI}$, $M_{DET}$, and $M_{CLS}$ remain exactly the same as DE-ViT. For details of these modules, please refer to DE-ViT~\cite{zhang2023detect}.
Architectures of our novel modules, i.e., $M_{LIF}$, $M_{IR}$, $M_{DP}$ are stated in Sec.~\ref{sec:method}. The MLP contained in the $M_{IR}$ module is a fully connected layer and a softmax.

\noindent\textbf{Implementation Details.}  
The mAPs for 1/5/10 shots are reported.
Besides the vanilla OVD methods that detect novel objects directly, we adopt the "pretrain, finetune, and test" pipeline for all evaluated methods as in Sec.~\ref{sec:trainingandinfer}. 
In the case of our CD-ViTO, the base DE-ViT model pretrained on COCO is taken, 
then the $M_{DET}$, $M_{CLS}$, $M_{LIF}$, $M_{IR}$, and $M_{DP}$ are tuned on novel support set $S$ using the loss as in Eq.~\ref{eq:loss}.
Finetuning details e.g., hyperparameters, optimizer, learning rate, and epochs will be attached in the supplementary materials.

\subsection{Evaluated Methods}\label{sec:exp-evaluatedmethods}

To conduct thorough experiments and offer valuable insights, we explore various methods across four types in the newly proposed CD-FSOD benchmark.

\noindent\textbf{Typical FSOD Methods.} 
Several FSOD methods, including  Meta-RCNN~\cite{yan2019meta}, TFA~\cite{wang2020frustratingly}, FSCE~\cite{sun2021fsce}, and DeFRCN~\cite{qiao2021defrcn}, are included to examine whether these traditional algorithms can handle the huge domain gap posed by CD-FSOD.

\noindent\textbf{CD-FSOD Methods.} 
Distill-cdfsod~\cite{xiong2023cd} and MoFSOD~\cite{lee2022rethinking} are two works for CD-FSOD. Distill-cdfsod is taken as a competitor. However,  
MoFSOD explores factors like finetuning and pretrained datasets, but doesn't propose a specific method, thus precluding numerical comparisons.

\noindent\textbf{ViT-based OD Methods.} Since most of the typical FSOD methods adopt the ResNet50~\cite{he2016deep} as the backbone, we would like to study how the large-scale pretrained ViT~\cite{dosovitskiy2020vit} models can benefit CD-FSOD. To that end, we adapt the ViTDeT~\cite{li2022exploring}, a flagship model that uses ViT for standard object detection, to our CD-FSOD. For clarity, we refer to the adapted ViTDeT as "ViTDeT-FT".

\noindent\textbf{Open-Set FSOD/OD Methods.} 
We study DE-ViT~\cite{zhang2023detect}, which excels in FSOD benchmarks. Considering the straightforward nature of finetuning, we also report results using the best finetuning strategy on DE-ViT, referred to as "DE-ViT-FT" (details in supplementary materials). Besides the purely vison-based DE-ViT, we evaluate another widely known vision-language matching OVD detector, namely Detic~\cite{zhou2022detecting}, and finetune it, further resulting in "Detic-FT".

\begin{table*}[ht]
\centering
\caption{Main Results (mAP) on CD-FSOD benchmark. Four types of object detection methods including typical FSOD, CD-FSOD, ViT-based OD, and open-set based OD/FSOD methods are evaluated and compared. The $\circ$ denotes results from Distill-cdfsod~\cite{xiong2023cd}; $\dagger$ denotes that the methods are developed by us or the results are reported by us. The best results are highlighted in blue. "Avg." means the average result. \label{tab:main}} 
\resizebox{0.95\columnwidth}{!}{
\begin{tabular}{cllcccccccc}
\toprule
& \textbf{Method} & \textbf{Backbone} & \textbf{ArTaxOr} & \textbf{Clipart1k} &  \textbf{DIOR} &   \textbf{DeepFish}   &  \textbf{NEU-DET}  & \textbf{UODD}  &  \textbf{Avg.} \\
\midrule
\multirow{7}{*}{\rotatebox{90}{1-shot}}

& Meta-RCNN $\circ$ ~\cite{yan2019meta} & ResNet50 & 2.8 & - & 7.8 & - & - & 3.6 & / \\

& TFA w/cos $\circ$~\cite{wang2020frustratingly} & ResNet50 & 3.1 & - & 8.0 & - & - & 4.4 & / \\

& FSCE $\circ$~\cite{sun2021fsce} & ResNet50 &  3.7 & - & 8.6 & - & - & 3.9 & / \\ 

& DeFRCN $\circ$~\cite{qiao2021defrcn} & ResNet50 & 3.6 & - & 9.3 & - & - & 4.5  & / \\

\cline{2-10}

& Distill-cdfsod $\circ$~\cite{xiong2023cd} & ResNet50 & 
5.1 & 7.6 & 10.5 & nan &  nan & \cellcolor{blue!15}5.9 & / \\

\cline{2-10}

& ViTDeT-FT$\dagger$ ~\cite{li2022exploring} & ViT-B/14 & 5.9 & 6.1 & 12.9 & 0.9 & 2.4 & 4.0 & 5.4\\

\cline{2-10}

& Detic$\dagger$~\cite{zhou2022detecting} & ViT-L/14 & 0.6& 11.4 & 0.1	& 0.9 &	0.0 & 0.0 & 2.2  \\
& Detic-FT$\dagger$ ~\cite{zhou2022detecting} & ViT-L/14 & 3.2 &	15.1	& 4.1	& 9.0 & 	\cellcolor{blue!15}3.8	 & 4.2 & 6.6  \\
& DE-ViT$\dagger$~\cite{zhang2023detect} & ViT-L/14 & 0.4	& 0.5& 2.7	&0.4 & 	0.4	 & 1.5 & 1.0  \\
& DE-ViT-FT$\dagger$~\cite{zhang2023detect} & ViT-L/14 &  10.5 &	13.0 & 14.7 & 19.3 & 0.6 & 2.4 & 10.1\\
\cline{2-10} 
& \textbf{CD-ViTO (ours)} & ViT-L/14 &  \cellcolor{blue!15}21.0 & \cellcolor{blue!15}17.7	& \cellcolor{blue!15}17.8 & \cellcolor{blue!15}20.3	& 3.6	& 3.1 & \cellcolor{blue!15}13.9 \\

 \midrule
 \multirow{7}{*}{\rotatebox{90}{5-shot}}

& Meta-RCNN $\circ$ ~\cite{yan2019meta} & ResNet50 & 8.5 & - & 17.7 & - & - & 8.8 & / \\

& TFA w/cos $\circ$~\cite{wang2020frustratingly} & ResNet50 & 8.8 & - & 18.1 & - & - & 8.7 & / \\

& FSCE $\circ$~\cite{sun2021fsce} & ResNet50 & 10.2 & - & 18.7 & - & - & 9.6 & / \\ 

& DeFRCN $\circ$~\cite{qiao2021defrcn} & ResNet50 &  9.9 & - & 18.9 & - & - & 9.9 & / \\

\cline{2-10}

& Distill-cdfsod $\circ$~\cite{xiong2023cd} & ResNet50 &  
12.5 & 23.3 & 19.1 & 15.5 &  \cellcolor{blue!15}16.0 &  \cellcolor{blue!15}12.2 &  16.4 
\\

\cline{2-10}

& ViTDeT-FT$\dagger$ ~\cite{li2022exploring} & ViT-B/14 &  
20.9 &   23.3 & 23.3 & 9.0 & 13.5 & 11.1 &  16.9  \\

\cline{2-10}

& Detic$\dagger$~\cite{zhou2022detecting} & ViT-L/14 & 
0.6	& 11.4	& 0.1	&  0.9 & 	0.0 & 	0.0 &  2.2 \\
& Detic-FT$\dagger$ ~\cite{zhou2022detecting} & ViT-L/14 & 8.7& 20.2	 & 12.1	& 14.3  & 14.1	& 10.4 &  13.3  \\
& DE-ViT$\dagger$~\cite{zhang2023detect} & ViT-L/14 & 10.1	& 5.5	& 7.8	& 2.5	& 1.5 & 3.1 &  5.1 \\
& DE-ViT-FT$\dagger$~\cite{zhang2023detect} & ViT-L/14 & 38.0 &	38.1 &	23.4 &	21.2&	7.8	 & 5.0 & 22.3\\
\cline{2-10}
& \textbf{CD-ViTO (ours)}  & ViT-L/14 &  \cellcolor{blue!15}47.9	&  \cellcolor{blue!15}41.1	&  \cellcolor{blue!15}26.9 &  \cellcolor{blue!15}22.3 & 11.4 & 6.8 & \cellcolor{blue!15}26.1   \\

 \midrule
\multirow{7}{*}{\rotatebox{90}{10-shot}} 

& Meta-RCNN $\circ$ ~\cite{yan2019meta} & ResNet50 & 14.0 & - & 20.6 & - & - & 11.2& /  \\

& TFA w/cos $\circ$~\cite{wang2020frustratingly} & ResNet50 &14.8 & - & 20.5 & - & - & 11.8 & / \\

& FSCE $\circ$~\cite{sun2021fsce} & ResNet50 & 15.9 & - & 21.9 & - & - & 12.0 & /\\ 

& DeFRCN $\circ$~\cite{qiao2021defrcn} & ResNet50 & 15.5 & - & 22.9 & - & - & 12.1 & / \\ 

\cline{2-10}

& Distill-cdfsod $\circ$~\cite{xiong2023cd} & ResNet50 &  
18.1 & 27.3 & 26.5 & 15.5 & \cellcolor{blue!15}21.1 & 14.5 &  20.5 \\ 
			
\cline{2-10}

& ViTDeT-FT$\dagger$ ~\cite{li2022exploring} & ViT-B/14 & 23.4 & 25.6 & 29.4 & 6.5 & 15.8 & \cellcolor{blue!15}15.6  &  19.4 \\ 
		
\cline{2-10}

& Detic$\dagger$~\cite{zhou2022detecting} & ViT-L/14 & 0.6 &	11.4	& 0.1	& 0.9 &  0.0	& 0.0 & 2.2  \\
& Detic-FT$\dagger$ ~\cite{zhou2022detecting} & ViT-L/14 & 12.0 &	22.3 & 15.4	& 17.9 & 16.8	& 14.4 &  16.5 \\
& DE-ViT$\dagger$~\cite{zhang2023detect} & ViT-L/14 & 9.2 & 11.0 & 8.4	& 2.1 &	1.8 & 	3.1&  5.9  \\
& DE-ViT-FT$\dagger$~\cite{zhang2023detect} & ViT-L/14 &  49.2	& 40.8	& 25.6 & 21.3 & 8.8	& 5.4 &	25.2  \\
\cline{2-10}
 &  \textbf{CD-ViTO (ours)} & ViT-L/14 &  \cellcolor{blue!15}60.5&  \cellcolor{blue!15}44.3	&  \cellcolor{blue!15}30.8 & \cellcolor{blue!15}22.3 & 12.8 & 7.0 & \cellcolor{blue!15}29.6 \\
\bottomrule
\end{tabular}}
\end{table*}

\subsection{Main Results on CD-FSOD}\label{sec:exp-cdfsod}
The results for 1/5/10 shot scenarios on six novel target datasets are summarized in Tab.~\ref{tab:main}. 
What can be apparently drawn from the table is the remarkable superiority of our CD-ViTO over the base DE-ViT across all target datasets. Moreover, it outperforms other competitors on the majority of these datasets.
For instance, under the 10-shot setting on ArTaxOr, CD-ViTO surpasses Meta-RCNN by 332.1\% (from 14.0 to 60.5), ViTDeT-FT by 158.5\% (from 23.4 to 60.5), and Detic-FT by 404.2\% (from 12.0 to 60.5). These results strongly highlight the effectiveness of our CD-ViTO method in the context of FSOD across domains. Beyond performance improvement, we also delve into the following aspects:

\noindent\ding{172} \textbf{Does the domain gap pose challenges for FSOD models?}
Yes, it does. In examining the results of typical FSOD methods, we observe relatively low performance on the target datasets. Consider DeFRCN, for instance, as reported in its paper~\cite{qiao2021defrcn}, which achieves 9.3 mAP on COCO's novel testing set under the 1-shot setting. However, on ArTaxOr and UODD, it drops significantly to 3.6 (a decrease of 61.3\%) and 4.5 (a decrease of 51.6\%), respectively. Even the Distill-cdfsod, specifically designed for CD-FSOD, fails to surpass 9.3 on the majority of target domain datasets. These observations indicate the importance of advancing CD-FSOD methods.

\noindent\ding{173}  \textbf{Are ViT-based models superior to ResNet-based ones? }
The introduction of ViTDeT-FT aims to assess whether the ViT backbone can strengthen the performance of CD-FSOD. The comparison between ViTDeT-FT and ResNet50-based methods reveals that the effectiveness depends on the specific target datasets. ViTDeT-FT shows strong performance on ArTaxOr, Clipart1k, and DIOR, but it is less effective on DeepFish, NEU-DET, and UODD. Despite the absence of clear patterns, it can be observed that the choice of backbone is not a decisive factor in constructing accurate CD-FSOD models.

\noindent\ding{174}  \textbf{Can open-set models directly address the challenges of CD-FSOD?}
Unfortunately, the results of Detic and DE-ViT clearly show that the answer is no. Despite being designed to recognize any novel category, open-set detectors still face challenges posed by domain gaps.
It is noteworthy that Detic performs inference using category names and the query image, essentially adopting a zero-shot approach. Consequently, the results for 1/5/10 shots remain the same. On the other hand, DE-ViT requires support instances to construct prototypes, leading to variations in results. These findings highlight the nontrivial nature of solving the CD-FSOD by directly applying existing open-set models.

\noindent\ding{175}  \textbf{Can open-set models be significantly improved?}
Absolutely! While the performance of vanilla open-set models may be initially low, we have demonstrated that it can be substantially enhanced. This conclusion becomes evident when comparing Detic-FT to Detic and our CD-ViTO to DE-ViT. The success of Detic-FT and CD-ViTO underscores the necessity of finetuning for adapting open-set models to CD-FSOD. 
Coupled with finetuning and our other novel components, CD-ViTO achieves very impressive results, transforming the initial shortcomings of open-set models into a very promising solution for CD-FSOD.

\noindent\ding{176}  \textbf{Do different open-set models exhibit distinct performances? }
Yes.  We delve into the performance of DE-ViT and Detic, two models that achieve open-set learning through entirely different approaches. DE-ViT relies solely on vision features, while Detic relies on the relationship between visual features and category text.
Upon analyzing the results of DE-ViT-FT and Detic-FT, a clear trend emerges. DE-ViT-FT demonstrates a decisive advantage on ArTaxOr, Clipart1k, DIOR, and DeepFish, but falls behind Detic-FT on NEU-DET and UODD. We observe that when the Instance Boundary (IB) degree of the target dataset is slight or moderate, the vision-based DE-ViT-FT performs better. Conversely, in scenarios where boundaries are significantly indefinable, indicating strong similarity between foreground and background, methods like Detic-FT, which leverages both visual and text information, prove more competent.

\noindent\ding{177} \textbf{How do style, inter-class variance (ICV), and indefinable boundaries (IB) impact the domain gap?} We meticulously analyze the influence of style, ICV, and IB on domain gaps based on experimental results and benchmark analysis in Sec.~\ref{sec:benchmark}.
1) Style, often considered strongly related to the domain, has a relatively smaller impact on CD-FSOD compared to ICV and IB. For instance, DIOR shares relatively similar ICV and IB with COCO but has a different style. Despite this, the performance of typical FSOD methods does not significantly drop on DIOR. For example, DeFRCN achieves 9.3 mAP on both COCO's testing set and DIOR under the 1-shot setting. Similarly, the cartoon-style Clipart1k, distinct from the photorealistic COCO, also receives relatively high mAPs.
2) ICV can cause domain issues but appears manageable. 
ArTaxOr, with the same style and similar IB as COCO, but with fine-grained categories (arthropods) compared to the coarse-level semantic concepts in COCO. This difference in ICV leads to a significant drop in performance for typical FSOD methods on ArTaxOr, such as from 9.1 to 3.6 on 1-shot using DeFRCN. 
However, we observe that this negative effect can be mitigated by utilizing better vision features, as demonstrated by ViTDeT-FT, DE-ViT-FT, and our CD-ViTO.
3) The change in IB poses the greatest challenge for CD-FSOD. Nearly all types of methods struggle to achieve satisfactory results on NEU-DET and UODD when significant IB exists. Blurred boundaries make it challenging to detect correct bounding boxes, resulting in poor overall performance.

\begin{figure}
\centering	{\includegraphics[width=0.9\linewidth]{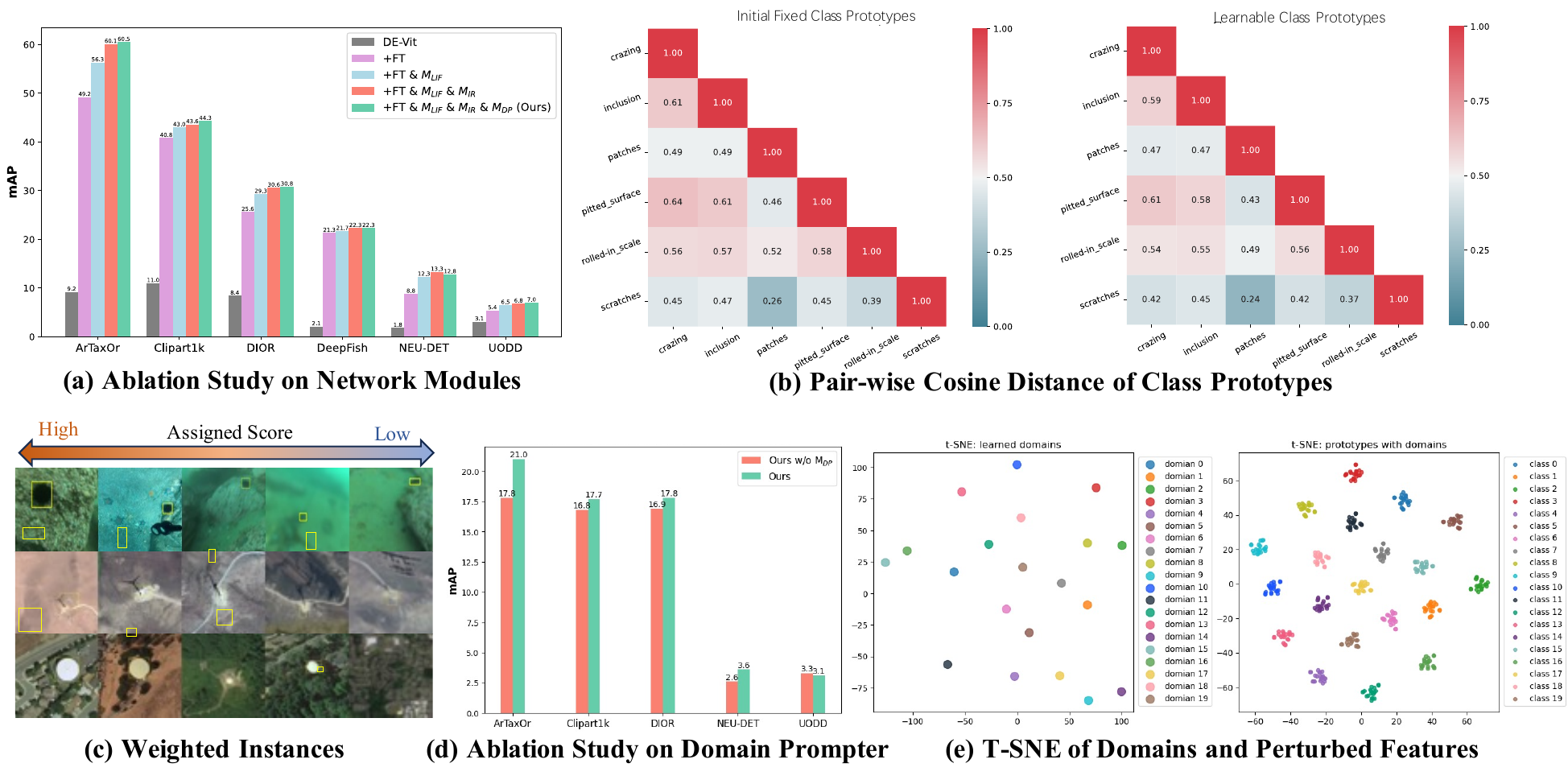}}
\caption{Ablation study on our proposed modules.}\label{fig:abla-module} 
\end{figure}

\subsection{Analysis on the Proposed Modules}\label{sec:5.3}
To demonstrate the effectiveness of our proposed components, we conduct ablation studies by incrementally adding them to the base DE-ViT. The 10-shot results are presented in Fig.~\ref{fig:abla-module}(a). 
Overall, we observe that: 1) By simply finetuning top heads, the performance is significantly improved. The results reinforce the observed phenomenon and analysis from Tab.~\ref{tab:main} that finetuning is crucial in CD-FSOD. 
2) The learnable instance features further boost performance clearly, indicating that we do make the features better by optimizing them. 
3) Although not as pronounced as in finetuning and $M_{LIF}$, the results still demonstrate the effectiveness of our $M_{IR}$ and $M_{DP}$ modules.

\noindent\textbf{More Discussions on Learnable Instance Features.} 
We propose the learnable instance features $M_{LIF}$ to tackle the small ICV issue hoping to make the visual features more discriminative. To verify whether the $M_{LIF}$ module meets expectations, we calculate the pair-wise cosine distances of class prototypes and compare the results of our learnable class prototypes produced by $M_{LIF}$ with that of the initial fixed ones. Results on 5-shot NEU-DET are given in Fig.~\ref{fig:abla-module}(b), we observe that the cosine distances between classes are consistently declining. This indicates that our $M_{LIF}$ enlarges the ICV.

\noindent\textbf{More Discussions on Instance Reweighting.} The instance reweighting module $M_{IR}$ is presented to assign higher weights to the instances with slight IB values. Thus, we visualize several weighted instances ordered by weights from high to low. 5-shot results from three datasets are demonstrated in Fig.~\ref{fig:abla-module}(c). Results show that instances with relatively clear boundaries are given higher scores; conversely, those with confusing boundaries receive lower scores.

\noindent\textbf{More Discussions on Domain Prompter.} 
The $M_{DP}$ is designed to generate virtual domains as useful perturbations. Different from the relatively marginal improvements in 10-shot (Fig.~\ref{fig:abla-module}(a)), the 1-shot results on Fig.~\ref{fig:abla-module}(d) show significant gains.
Note that we don't apply $M_{DP}$ on DeepFish since it only contains one object category and the slight drop in UODD indicates that training the new parameters introduced by $M_{DP}$ in extremely data-limited cases (3 classes for UODD) may pose challenges.
Besides quantified results, in Fig.~\ref{fig:abla-module}(e), we provide the t-SNE of our learned domains and the perturbed features by adding domains to the class prototypes. The 1-shot results on Clipart1K are visualized. The results reveal that: 1) the learned domains exhibit diversity; 2) adding domains to the class prototypes generates new data but keeps the semantic concepts maintained. 
Finally, we term these learnable perturbations as "domain" prompters rather than "style" prompters since learning meaningful styles without explicit style information is challenging. 
The term "domain" is used broadly to encompass factors, including style, that could induce changes in data distribution.

\section{Conclusion}
In conclusion, this paper presents several key contributions to the field of cross-domain few-shot object detection (CD-FSOD). Firstly, we meticulously reviewed and reorganized object detection datasets, establishing a novel benchmark specifically tailored for CD-FSOD. Secondly, we conducted a comprehensive study involving various methods, including typical FSOD models, CD-FSOD models, ViT-based OD models, and open-set-based OD/FSOD models, based on this benchmark. Through careful analysis, we provided detailed insights, marking the first exploration of open-set based detectors for CD-FSOD. Thirdly, leveraging the state-of-the-art DE-ViT, we introduced the novel CD-ViTO method, enhancing the open-set detector with our novel techniques, i.e., finetuning, learnable instance features, instance reweighting, and domain prompter. The effectiveness of CD-ViTO and its modules are thoroughly validated on target domains. We believe our contributions will propel advancements for CD-FSOD.

\bibliographystyle{splncs04}
\bibliography{main}

 \newpage

\begin{spacing}{1.2}
\begin{center}
\textbf{\large Supplemental Materials: Cross-Domain Few-Shot Object Detection via Enhanced Open-Set Object Detector} 
\end{center}
\end{spacing}

We first provide the implementation details in  Sec.~\ref{sec:supp-imple}, and then show more experimental results of our proposed methods in Sec.~\ref{sec:supp-results}. 
Finally, in Sec.~\ref{sec:supp-metric}, more details about our proposed benchmark are provided.

\section{Implementation Details}\label{sec:supp-imple}
We implement six methods for CD-FSOD as in Sec.~\ref{sec:exp-evaluatedmethods}, including ``ViTDeT-FT'' adapted from ViTDeT~\cite{li2022exploring}, Detic~\cite{zhou2022detecting}, ``Detic-FT'' adapted from Detic, DE-ViT~\cite{zhang2023detect}, ``DE-ViT-FT'' from DE-ViT, and our ``CD-ViTO''. 
For all the models, as stated in Sec.~\ref{sec:trainingandinfer}, we follow the ``pretrain, finetuning, and testing" pipeline, and the models pretrained on COCO are taken directly as the result of the pretrain stage. The finetuning and testing details are as follows. 

\paragraph{\textbf{ViTDeT-FT:}} We finetune the whole ViTDeT using target support images as training data. Specifically, the AdamW is used as the optimizer, the tuning epoch is set as 100. As for the learning rate, we initially set it as 1.0 and as the finetuning progressed, the learning rate is first reduced to 0.1, and finally to 0.01. The fine-tuned model is used for inference directly.

\paragraph{\textbf{Detic:}} Detic is an open-vocabulary based object detection method that can detect any novel object without finetuning, thus we skip the finetuning step and perform the testing stage directly. 

\paragraph{\textbf{Detic-FT:}} Detic-FT is the same as the vanilla Detic except that we finetune it on target support images. The SGD optimizer with a learning rate of 0.02 is used for optimizing the network.

\paragraph{\textbf{DE-ViT:}}  
Despite not using category information, DE-ViT builds a powerful open-set detector for arbitrary objects and achieves the SOTA performance in both the open-vocabulary setting and the few-shot setting.
Thus, we use the DE-ViT model well-trained on the COCO dataset to perform direct inference on the CD-FSOD benchmark. Note that the hyperparameter Top-K ($K$) in DE-ViT has a large impact on the results. Experimentally, we find that setting $K$ to 1 produces better results on the CD-FSOD benchmark when inference is done directly without finetuning. Consequently, we set $K$ to 1, while maintaining all the other settings consistent with the vanilla DE-ViT configuration.

\paragraph{\textbf{DE-ViT-FT:}}  
We build DE-ViT-FT by finetuning the $M_{DET}$, $M_{CLS}$ of DE-ViT (as studied in Tab.~\ref{tab:abla-finetune}). All the other implementation details e.g., optimizer, and learning rate are the same as our CD-ViTO. 

\paragraph{\textbf{CD-ViTO:}}
Based on DE-ViT, we develop CD-ViTO as in Sec.~\ref{sec:method}. 
The hyperparameter $N_{bg}$, $\alpha$ ratio of $M_{IR}$ module, $\tau$ temperature for $\mathcal{L}_{proto}$, $\tau$ temperature for $\mathcal{L}_{domain}$  are set as 530, 0.7, 2, 0.1 for all the target datasets. While the value $N_{dom}$ means the number of virtual domains depending on the number of target classes $N$, specifically, we have $N_{dom} = 2*N$.  The hyperparameter Top-K ($K$) in DE-ViT is set as 5. For datasets with number of classes $N$ less than 5, we have $K=N$. We tune the trainable parameters on 1-shot around 80 epochs, and on 5/10-shot around 40 epochs. The SGD with the learning rate as 0.002 is used as the optimizer. Experiments are performed on four RTX3090 GPUs.

\section{More Results}\label{sec:supp-results}
\subsection{Results on Standard FSOD}
We also conduct experiments on standard FSOD to see if our proposed method could further improve the base DE-ViT~\cite{zhang2023detect} on in-domain datasets. 
For a fair comparison, the hyperparameter $K$ is set to 10 as in DE-ViT~\cite{zhang2023detect} while the learning rate is selected from $\{0.0002, 0.00005\}$, and the epoch is set to about 100. The value of $N_{bg}$, $\alpha$, $\tau$ for $\mathcal{L}_{proto}$, $\tau$ for $\mathcal{L}_{domain}$ is set to 530, 0.7, 2, 0.1, respectively. Experiments on standard FSOD are performed on one A800 GPU.

\begin{table*}[h!]\scriptsize
\vspace{-0.2in}
\caption{
Results (nAP, nAP50, and nAP75) on COCO FSOD benchmark. The nAP denotes mAP for novel classes. The best results are highlighted in blue.
\label{tab:coco-fewshot}}
\vspace{-0.1in}
\scriptsize \centering 
\begin{tabular}{lc|ccc|ccc}\toprule
 \multicolumn{2}{c}{\multirow{2}{*}{\textbf{Method}}} & \multicolumn{3}{c}{\normalfont \textbf{10-shot}} & \multicolumn{3}{c}{\normalfont \textbf{30-shot}} \\
\cmidrule{3-8}
 & &nAP &nAP50 &nAP75 &nAP &nAP50 &nAP75 \\\midrule
\multicolumn{2}{l|}{FSRW \citep{kang2019few}} &5.6 &12.3 &4.6  &9.1 &19 &7.6 \\
\multicolumn{2}{l|}{Meta R-CNN \citep{yan2019meta}} &6.1 &19.1 &6.6 &9.9 &25.3 &10.8 \\
\multicolumn{2}{l|}{TFA \citep{wang2020frustratingly}} &10 &19.2 &9.2  &13.5 &24.9 &13.2 \\
\multicolumn{2}{l|}{FSCE \citep{sun2021fsce}}  &11.9 &- &10.5 &16.4 &- &16.2 \\
\multicolumn{2}{l|}{Retentive RCNN \citep{fan2021generalized}} &10.5 &19.5 &9.3  &13.8 &22.9 &13.8 \\
\multicolumn{2}{l|}{HeteroGraph \citep{han2021query}} &11.6 &23.9 &9.8 & 16.5 &31.9 &15.5 \\
\multicolumn{2}{l|}{Meta Faster R-CNN \citep{han2022meta}} &12.7 &25.7 &10.8 &16.6 &31.8 &15.8 \\
\multicolumn{2}{l|}{LVC \citep{kaul2022label}}  &19 &34.1 &19 &26.8 &45.8 &27.5 \\
\multicolumn{2}{l|}{Cross-Transformer \citep{han2022few}}  &17.1 &30.2 &17 &21.4 &35.5 &22.1 \\
\multicolumn{2}{l|}{NIFF \citep{guirguis2023niff}}  &18.8 &- &- &20.9 &- &- \\
\multicolumn{2}{l|}{DiGeo \citep{ma2023digeo}} &10.3 &18.7 &9.9  &14.2 &26.2 &14.8 \\\midrule
\multirow{3}{*}{DE-ViT~\cite{zhang2023detect}} &ViT-S/14 & 27.1 & 43.1 & 28.5 &26.9 & 43.1 & 28.4 \\
&ViT-B/14 & 33.2 & 51.4 & \cellcolor{blue!15}35.5 & 33.4 & 51.4 & 35.7 \\
&ViT-L/14 & 34.0 & 53.0 & 37.0 & 34.0 & 52.9 & 37.2 \\

\hline
\multirow{3}{*}{\textbf{CD-ViTO (ours)}} & ViT-S/14 &   \cellcolor{blue!15}27.5 & 	\cellcolor{blue!15}43.8 & \cellcolor{blue!15}28.8 & \cellcolor{blue!15}27.8 & \cellcolor{blue!15}43.5 & \cellcolor{blue!15}29.5\\
&ViT-B/14  & \cellcolor{blue!15}33.4 & \cellcolor{blue!15}51.8 & \cellcolor{blue!15}35.5 & \cellcolor{blue!15}33.6 & \cellcolor{blue!15}51.6 & \cellcolor{blue!15}36.1\\
&ViT-L/14 & \cellcolor{blue!15}35.3& \cellcolor{blue!15}54.9	&  \cellcolor{blue!15}37.2	&  \cellcolor{blue!15}35.9 & \cellcolor{blue!15}54.5	& \cellcolor{blue!15}38.0 \\
\bottomrule
\vspace{-0.3in}
\end{tabular}
\end{table*}

Results are summarized in Tab.~\ref{tab:coco-fewshot}, as in previous works~\cite{zhang2023detect, wang2020frustratingly, zhang2023detect}, the nAP (mAP for target data), nAP50, and nAP75 for 10/30 shots are reported. From the results, we notice that with our CD-ViTO successfully improves the performance of the base DE-ViT on all different ViT backbones and both 10 and 30 shots. This makes us build a new SOTA for FSOD.  We attribute the success primarily to 1) even sets partitioned from the same dataset may have some distribution shift issues; 2) with the learnable instance features our prototypes are more likely to approach the true class centers compared to fixed ones.

\subsection{Ablation Results on Each Module}
\paragraph{\textbf{Ablation on Different Backbones.}} In Tab.~\ref{tab:main}, we report the results of our CD-ViTO employing ViT-L/14 as the backbone, and we further provide the results of building our method upon the ViT-S/14 and ViT-B/14 backbones. As shown in Tab.~\ref{tab:vit}, the results generally indicate that ViT-L/14 performs the best, followed by ViT-B/14 and then ViT-S/14, which aligns with the expectation that ViT-L/14 embeds features more effectively. 

\begin{table*}[h] \scriptsize
\vspace{-0.2in}
\caption{The 1/5/10-shot results of CD-ViTO with different ViT backbones.}\label{tab:vit}
\vspace{-0.1in}
\centering
\begin{tabular}{clcccccccc}
\toprule
& \textbf{Backbone} 
& \textbf{ArTaxOr} & \textbf{Clipart1k} & \textbf{DIOR} &  \textbf{DeepFish}   & \textbf{NEU-DET}  & \textbf{UODD}  & \textbf{avg} \\
\midrule
\multirow{3}{*}{\rotatebox{90}{1-shot}}
  & ViT-S/14 & 16.4& 7.8 & 14.0 & 3.3	& 3.8& 2.6 & 8.0 \\
 & ViT-B/14 & 13.6  & 13.6	    & 15.7	& 13.2& \textbf{4.0}	                 & 2.4 & 10.4 \\
 & ViT-L/14 & \textbf{21.0}	& \textbf{17.7} & \textbf{17.8} & \textbf{20.3} & 3.6	& \textbf{3.1}	& \textbf{13.9} \\

 \midrule
 \multirow{3}{*}{\rotatebox{90}{5-shot}} 

 & ViT-S/14 & 36.2	& 25.5	& 20.4	& 20.4	& 10.5	& 6.1	& 19.9 \\
 & ViT-B/14 & 43.1& 36.3	& 22.8	& 20.2	& 10.7	& \textbf{7.4}	& 23.4 \\
 & ViT-L/14 & \textbf{47.9} & \textbf{41.1}	& \textbf{26.9}	& \textbf{22.3}	& \textbf{11.4} & 6.8 & \textbf{26.1} \\

 \midrule
\multirow{3}{*}{\rotatebox{90}{10-shot}}

 & ViT-S/14 & 42.5 & 29.8 & 23.6	& 21.2	& 12.8 & 5.9 & 22.6 \\
  & ViT-B/14 & 54.8  & 40.4	& 27.4& 19.5	& \textbf{13.1}	&6.2& 26.9\\
 & ViT-L/14 & \textbf{60.5}	& \textbf{44.3} & 	\textbf{30.8} &	\textbf{22.3}	& 12.8 & 	\textbf{7.0} & 	\textbf{29.6} \\

\bottomrule
\end{tabular}
\vspace{-0.25in}
\end{table*}

\paragraph{\textbf{Ablation on Effectiveness of Modules.}} 
Recall that in Fig.~\ref{fig:abla-module}(a) in the main file, we demonstrate the effectiveness of modules by adding our new proposed modules one by one. To further show the effectiveness of our modules, we further conduct ablation studies that add each of the modules separately into DE-ViT-FT. The 10-shot results are shown on Tab.~\ref{tab:supp-abla-m}.

Results show that: 1) Basically, all the modules work well improving the DE-ViT-FT clearly. 2) Overall, we observe that $M_{LIF}$ improves the result at most, then comes the $M_{IR}$ and $M_{DP}$. 3) With all modules added, our full CD-ViTO model achieves the best results on all target datasets. The conclusions are consistent with the main file.

\begin{table}[h!]\scriptsize
    \centering
    \vspace{-0.15in}
    \caption{The 10-shot results of ablation study on modules.}\label{tab:vit}
    \vspace{-0.1in}
    \resizebox{0.85\linewidth}{!}{
    \begin{tabular}{c|c|c|c|c|c|c} \hline 
         &  \textbf{ArTaxOr}&  \textbf{Clipart1k}&  \textbf{DIOR}&  \textbf{FISH}&  \textbf{NEU-DET}& \textbf{UODD}\\ \hline 
         DE-ViT-FT& 49.2 & 40.8 & 25.6 & 21.3 & 8.8 & 5.4 \\ \hline
         \textbf{+$M_{LIF}$}&  56.3&  43.0&  29.3&  21.7&  12.3& 6.5\\ \hline 
         \textbf{+$M_{IR}$}&  56.8&  42.8&  28.5&  21.8&  12.4& 5.7\\ \hline 
         \textbf{+$M_{DP}$}&  50.1 & 42.8  & 26.5 & 22.2 &  9.9 & 5.6      \\ \hline 
         + all (\textbf{CD-ViTO, ours})& \textbf{60.5} & \textbf{44.3} & \textbf{30.8} &  \textbf{22.3} & \textbf{12.8}  & \textbf{7.0} \\ \hline
    \end{tabular}   }
        \vspace{-0.25in}
    \label{tab:supp-abla-m}
\end{table}

\paragraph{\textbf{Ablation on Finetuning.}} 
In this paper, we reveal that finetuning plays a key role in applying the open-set detector for CD-FSOD. To further identify the optimal finetuning strategy, we conduct an extensive ablation study on the main basic modules of DE-ViT: pretrained DINOv2, $M_{RPN}$, $M_{DET}$, and $M_{CLS}$ with or without our new modules. Results are reported in Tab.~\ref{tab:abla-finetune}, note we use $M_{novel}$ to represent our $M_{LIF}$, $M_{IR}$, and $M_{DP}$ modules. 

We highlight several points. 
1) overall, finetuning two heads already could achieve impressive results compared to the base DE-ViT which indicates the effectiveness of finetuning key modules in the context of CD-FSOD.
2) Equipped with our novel modules, the performance on various target datasets is further improved steadily. 
3) Finetuning the $M_{RPN}$ on the base of ours may further improve results marginally, however, it would bring much more learnable parameters and thus not adopted.
4) Results also suggest that it is hard to optimize the well-trained DINOv2 backbone with few examples. Therefore, freeing the DINO backbone is not advisable. 
Based on the result, we set "FT two heads" as our "DE-ViT-FT". 
Additionally, from the number of trainable parameters (No.TrainP), it is easy to tell that our novel modules are highly efficient.

\begin{table*}[h!]\small
\vspace{-0.15in}
\caption{The 10-shot ablation results on finetuning based on ViT-L/14. } \label{tab:abla-finetune}
\vspace{-0.15in}
\centering
\resizebox{1.\columnwidth}{!}{
\begin{tabular}{lcccccccccccl}
\toprule
 \textbf{Method} &  \textbf{$M_{DET}$} & \textbf{$M_{CLS}$}  & \textbf{$M_{novel}$}  & \textbf{$M_{RPN}$} & \textbf{$DINO$} & \textbf{ArTaxOr} & \textbf{Clipart1k} & \textbf{DIOR} & \textbf{DeepFish} & \textbf{NEU-DET} & \textbf{UODD} & \textbf{No.TrainP}\\
\midrule
Base DE-ViT & - & - & - & - & - & 9.2 & 11.0 & 8.4 & 2.1 & 1.8 & 3.1 & None\\
FT two heads & \checkmark & \checkmark & - & - & - & 49.2 & 40.8 & 25.6 & 21.3 & 8.8 & 5.4 & + 7.9M\\
Ours & \checkmark & \checkmark & \checkmark & - & - & \textbf{60.5}	& 44.3 & 	30.8 &	\textbf{22.3}	& \textbf{12.8} & 7.0  & + \textbf{0.8M} \\
Ours + FT RPN &  \checkmark & \checkmark & \checkmark & \checkmark & - & 58.8	& \textbf{44.7} & \textbf{31.1}	& \textbf{22.3}	& 12.6	& \textbf{7.7} & + 18.1M\\
Ours + FT RPN + DINO & \checkmark & \checkmark & \checkmark & \checkmark & \checkmark  & 0.3 & 0.1 &	0.1 & 	2.2	& 1.7 & 1.3 & + 304.4M \\
\bottomrule
\vspace{-0.3in}
\end{tabular}}
\end{table*}

\paragraph{\textbf{Ablation on Learnable Instance Features $M_{LIF}$.}} We conduct experiments to evaluate how different strategies of $M_{LIF}$ will affect our results. Concretely, compared to our CD-ViTO that sets both the foreground object instances $F_{ins}^{ob}$ and background instances $F_{ins}^{bg}$ trainable, we also study the results of only setting the $F_{ins}^{ob}$ learnable and the $F_{ins}^{bg}$ learnable. The results of the 10-shot are presented in Tab.~\ref{tab:abla-learnPro}. The finetuning and the $M_{LIF}$ are applied.
These results illustrate that optimizing both foreground and background instances is the best in most cases.

\begin{table*}\scriptsize
\vspace{-0.15in}
\caption{The 10-shot ablation results on $M_{LIF}$ based on ViT-L/14.}\label{tab:abla-learnPro}
\vspace{-0.1in}
\centering
\begin{tabular}{llcccccc}
\toprule
 \textbf{$F_{ins}^{ob}$} & \textbf{$F_{ins}^{bg}$} & \textbf{ArTaxOr} & \textbf{Clipart1k} 
& \textbf{DIOR} & \textbf{DeepFish} & \textbf{NEU-DET} & \textbf{UODD} \\
\midrule
\checkmark & -  & 53.1	& \textbf{43.4}& 26.9& \textbf{22.1}	& 10.0 & 5.5\\
- & \checkmark & 54.5& 42.0	& 28.9	& 21.7	& 11.9	& 6.0 \\
\checkmark & \checkmark  & \textbf{56.3} & 43.0    & \textbf{29.3}	& 21.7	& \textbf{12.3}	& \textbf{6.5} \\
\bottomrule
\end{tabular}
\vspace{-0.25in}
\end{table*}

\paragraph{\textbf{Ablation on Instance Reweighting Module $M_{IR}$.}} Recall that we set a hyperparameter $\alpha$ for fusing the deformed prototype features $fc(F_{pro}^{att})$ and the averaged prototype features $F_{pro}^{avg}$ i.e., $\alpha fc(F_{pro}^{att}) + (1-\alpha)F_{pro}^{avg}$.  Thus, we conduct an ablation study on the value of $\alpha$. Results are shown in Tab.~\ref{tab:abla-att}. The finetuning, $M_{LIF}$, and $M_{IR}$ are applied. According to the results, we set $\alpha$ as 0.7.

\begin{table*}\scriptsize
\vspace{-0.15in}
\caption{The 10-shot ablation results on $M_{IR}$ based on ViT-L/14.}\label{tab:abla-att}
\vspace{-0.1in}
\centering
\begin{tabular}{ccccccc}
\toprule
 \textbf{$\alpha$} & \textbf{ArTaxOr} & \textbf{Clipart1k} 
& \textbf{DIOR} & \textbf{DeepFish} & \textbf{NEU-DET} & \textbf{UODD} \\
\midrule
0.1	& 46.3 & 40.5 & 26.0 &	21.4 & 10.6 & 5.8 \\
0.3	& 47.4 & 40.2 & 26.4 & 	21.4 & 11.0 & 6.0 \\
0.5 & \textbf{47.9} & \textbf{41.9} &	26.9 &	21.9 & 11.3	& 6.5 \\
0.7 (ours)	& 47.3 & 41.4 &	\textbf{27.3} &	22.3 & \textbf{12.0} & 6.8 \\
0.9	& 32.8 & 37.9 &	24.2 & 	\textbf{22.6} & 8.9 & \textbf{7.0} \\
\bottomrule
\end{tabular}
\vspace{-0.25in}
\end{table*}

\paragraph{\textbf{Ablation on Domain Prompter $M_{DP}$.}}  
our $M_{DP}$ introduces a total of $N_{dom}$ learnable latent vectors as virtual domains.  Therefore, we conduct ablation studies on the values of $N_{dom}$. The 10-shot results are summarized in Tab.~\ref{tab:abla-dp}. All the finetuning, the $M_{LIF}$, the $M_{IR}$, and the $M_{DP}$ are adopted. As stated in Sec.~\ref{sec:5.3}, $M_{DP}$ is not applied to the DeepFish cause it only contains a single class. 
From the results, we observe that: 
with $N_{dom} = 2*N$, we achieve the best results. 
We establish a positive correlation between $N_{dom}$ and the number of classes $N$ through our empirical observation, indicating that a greater number of trainable parameters is permissible when there are more available training instances. 

\begin{table*}\scriptsize
\vspace{-0.15in}
\caption{The 10-shot ablation results on $M_{DP}$ based on ViT-L/14.} \label{tab:abla-dp}
\vspace{-0.1in}
\centering
\begin{tabular}{lcccccc}
\toprule
 \textbf{$N_{dom}$} & \textbf{ArTaxOr} & \textbf{Clipart1k} 
& \textbf{DIOR} & \textbf{DeepFish} & \textbf{NEU-DET} & \textbf{UODD} \\
\midrule
$1*N$ & 59.8 & 44.2 & 30.2 & - & \textbf{12.9} & 6.2 \\
$2*N$ (ours) & \textbf{60.5} & 	44.3 &	\textbf{30.8} &	- & 12.8 &	\textbf{7.0} \\
$3*N$  & 59.9 & 	\textbf{44.8} & 30.6	 & - & 12.8 &	6.7 \\
\bottomrule
\end{tabular}
\vspace{-0.25in}
\end{table*}

\paragraph{\textbf{Time Complexity and FLOPs.}}  
To further investigate the additional cost caused by our modules, we separately add modules on DE-ViT-FT for inference and report the time complexity and FLOPs on 10-shot as in Tab.~\ref{tab:supp-cost}. From the results, we highlight that: 1) both $M_{LIF}$ and $M_{DP}$ don't cause any cost since $M_{LIF}$ only change the feature while $M_{DP}$ is solely used during the finetuning stage. 2) The $M_{IR}$ causes minimal time and FLOPs. 

\begin{table}[h]\scriptsize
\centering
\vspace{-0.15in}
\caption{The time complexity and FLOPs of the modules on 10-shot. }
\vspace{-0.1in}
\resizebox{0.9
\linewidth}{!}{
\begin{tabular}{c|c|c|c}
\hline
& +$M_{LIF}$  & + $M_{IR}$  & + $M_{DP}$   \\ \hline
\textbf{Time/Query} & \multicolumn{1}{c|}{\multirow{2}{*}{ +0 (\textbf{only change features})}} & \multicolumn{1}{c|}{+0.0006 S} & \multicolumn{1}{c}{\multirow{2}{*}{+0 (\textbf{only for finetuning})}} \\ \cline{1-1} \cline{3-3}
\textbf{FLOPs}      & \multicolumn{1}{c|}{}                                           & \multicolumn{1}{c|}{+14.1 M}   & \multicolumn{1}{c}{}                                          \\ \hline
\end{tabular}}\label{tab:supp-cost}
\vspace{-0.3in}
\end{table}

\subsection{More Visualization Results}

\begin{figure*}[h!]
\centering	{\includegraphics[width=.85\linewidth]{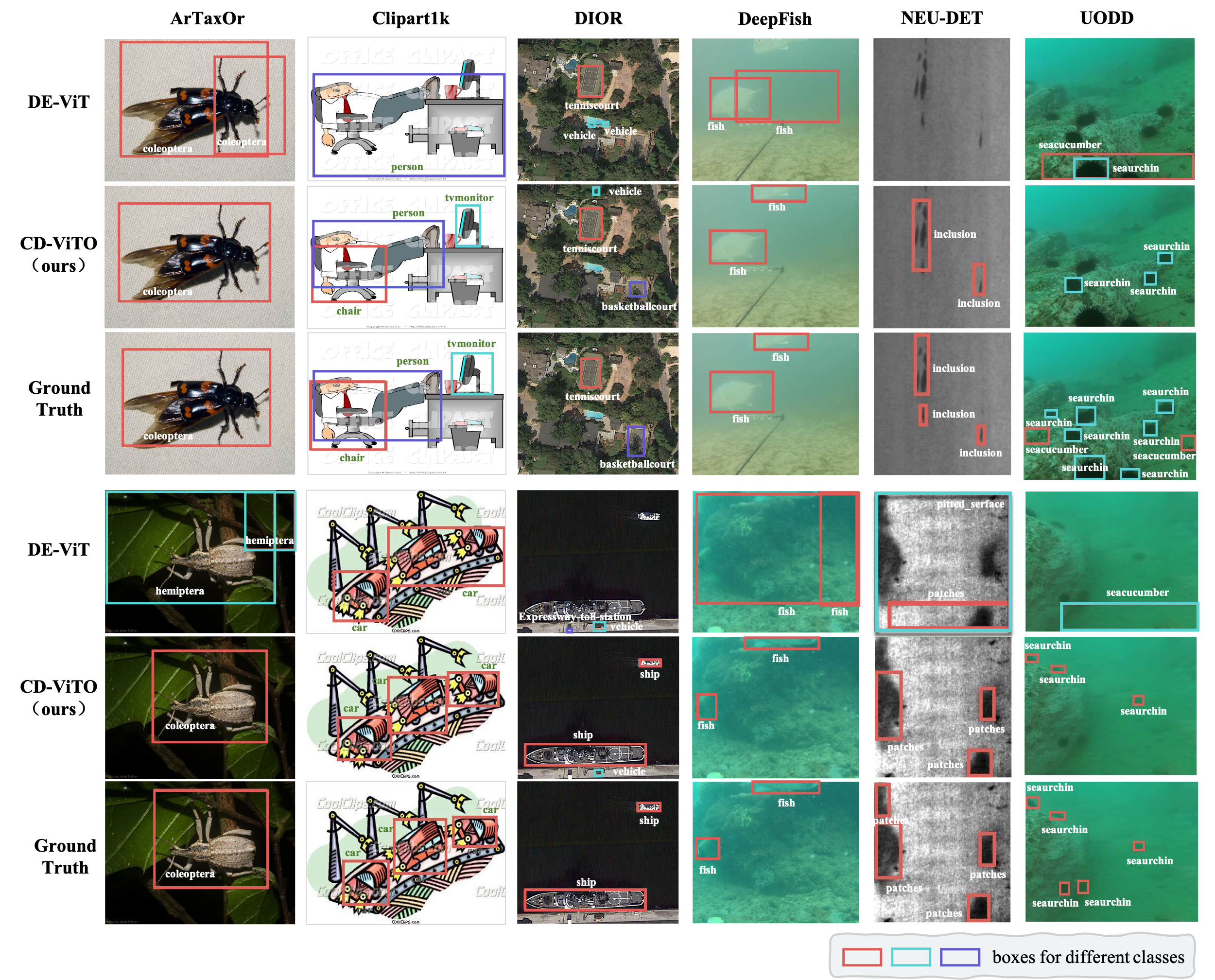}}
\vspace{-0.1in}
\caption{Visualization results of DE-ViT, our CD-ViTO, and ground truth. 
\label{fig:visulization} 
}
\vspace{-0.25in}
\end{figure*}

We also visualize several examples of our proposed CD-ViTO detector. 
As in Fig.~\ref{fig:visulization}, we compare the localization and classification results of CD-ViTO against the vanilla DE-ViT and ground truth. For each target dataset, two examples are given. We use different colors to denote various classes and use the text along with the box to represent the corresponding class category.

We observe that: 
1) The vanilla DE-ViT fails to identify the target objects. For example, the wrong areas are detected on ArTaxOr, meaningless large boxes are predicted on Clipart1k and DeepFish, and objects are ignored on DIOR, NEU-DET, and UODD. 
In addition, as revealed in the second example of ArTaxOr, even when the bounding box is close to the ground truth, DE-ViT will misclassify the category. 
These phenomenons indicate that the vanilla DE-ViT struggles to conduct both the localization and classification tasks on novel target domains; 
2) Our CD-ViTO largely improves the vanilla DE-ViT with more accurate localization and classification results observed. Generally, the predictions of our CD-ViTO align well with the ground truth, especially in the cases of  ArTaxOr, Clipart1k, and DeepFish. This indicates that we do enhance the vanilla DE-ViT obviously.
3) We also find there are some failure cases of our CD-ViTO in Fig.~\ref{fig:visulization}. For the NEU-DET and UODD, our CD-VITO also makes mistakes, such as missing some target objects. 
As analyzed in Sec.~\ref{sec:exp-cdfsod}, when the target datasets have significant IB, our method shows disadvantages. How to improve performance under confusing backgrounds would be one of our further works.

\subsection{More Discussion with Related Works}
Though discussed in related work, since AcroFOD~\cite{gao2022acrofod}, AsyFOD~\cite{gao2023asyfod}, MoFSOD~\cite{lee2022rethinking}, and Distill-cdfsod~\cite{xiong2023cd} all address few-shot object detection across domains. We further clarify the differences below.

Typically, 
1) AcroFOD and AsyFOD assume the source and target share the same class set, differing from our task essentially, which addresses entirely different class sets.
2) MoFSOD proposes a comprehensive benchmark but lacks a specific method. Additionally, the evaluation protocols differ: MoFSOD uses natural K-shot sampling, which samples K images per class regardless of the number of object instances per image. In contrast, we use balanced sampling, ensuring exactly K object instances per class. Our manner ensures the support instances are always few-shot and align with typical FSOD practices.
3) Distill-cdfsod is the most similar work. We inherit its three target datasets and evaluation method but introduce significant novel contributions. Compared to Distill-cdfsod, we propose new metrics, explore open-set detectors, and achieve much better results.
The above comparisons are summarized in Tab.~\ref{tab:supp-com}.

\begin{table}[h!]\scriptsize
\centering
\vspace{-0.15in}
\caption{More comparison with existing works.}
\vspace{-0.1in}
\resizebox{0.9\linewidth}{!}{
\begin{tabular}{c|l|l|l|l|l}
\hline
 & \textbf{Task}& \textbf{Protocol}& \textbf{Method}& \textbf{Metric}& \textbf{SOTA}\\ \hline
AcroFOD~\cite{gao2022acrofod}, AsyFOD~\cite{gao2023asyfod} & $\mathcal{C}_{S} = \mathcal{C}_{T}$         &        -      &   -        &     -    & -  \\ \hline
MoFSOD~\cite{lee2022rethinking} & $\mathcal{C}_{S} \cap \mathcal{C}_{T}=\emptyset$        &       natural K-shot           &    $\times$    & $\times$  &  - \\ \hline
Distill-cdfsod~\cite{xiong2023cd} & $\mathcal{C}_{S} \cap \mathcal{C}_{T} = \emptyset $            &    balanced K-shot          &     Distill-cdfsod      &        $\times$  & $\times$ \\ \hline
\textbf{CD-ViTO (ours)}&\cellcolor{blue!15}$\mathcal{C}_{S} \cap \mathcal{C}_{T} = \emptyset $              &     \cellcolor{blue!15}balanced K-shot        &    \cellcolor{blue!15}CD-ViTO    &   \cellcolor{blue!15}ICV, IB, style  & \cellcolor{blue!15}\checkmark      \\ \hline
\end{tabular}\label{tab:supp-com}
}
\vspace{-0.25in}
\end{table}

\section{More Details of Proposed Benchmark}\label{sec:supp-metric}
\subsection{Target Datasets}

The summarized information of the target datasets is given in Tab.~\ref{tab: target_data}. We also give brief descriptions of the target datasets as follows.

\noindent\textbf{ArTaxOr}~\cite{GeirArTaxOr}: A dataset about arthropods, including classes like insects, spiders, crustaceans, and millipedes.

\noindent\textbf{Clipart1k}~\cite{inoue2018cross}: This dataset features cartoon-like styles, making it well-suited for CD-FSOD. 
Coarse categories e.g., common indoor objects are included. 

\noindent\textbf{DIOR}~\cite{li2020object}:
Created for remote sensing object detection, DIOR contains a diverse set of annotated satellite objects like airplanes, ships, and golf fields.

\noindent\textbf{DeepFish}~\cite{saleh2020realistic}: This dataset offers high-resolution DeepFish images collected from different natural habitats.

\noindent\textbf{NEU-DET}~\cite{song2013noise}: Providing data for industrial surface defection, NEU-DET gathers instances of six typical surface defects (rolled-in scale, crazing, pitted surface, patches, inclusion, and scratches) of hot-rolled steel strips.

\noindent\textbf{UODD}~\cite{jiang2021underwater}: Collected underwater, UODD includes categories like sea urchins, sea cucumbers, and scallops. 

\begin{table*}[h!] 
\vspace{-0.22in}
  \caption{Information about the target datasets. "No." represents the number. }  \label{tab: target_data}
\vspace{-0.13in}
  \scriptsize
  \centering
  \begin{tabular}{llllccccccc}
    \toprule
     \textbf{Dataset} & \textbf{No. class} & \textbf{No. image}  & \textbf{No. box} & \textbf{Style} & \textbf{ICV} & \textbf{IB} \\
    \midrule
    ArTaxOr & 7 &  1383 & 1628 & photorealistic & small & slight \\
    Clipart1k & 20 & 500 & 1526 & cartoon & large & slight  \\
    DIOR & 20 & 5000 &28810 & aerial & medium & slight \\
    DeepFish & 1 & 909 &3029 & underwater & / & moderate  \\
    NEU-DET & 6 & 360 &	834	& industry & large & significant \\
    UODD & 3 & 506  & 3218 & underwater & small & significant \\
    \bottomrule
  \end{tabular}
\vspace{-0.4in}
\end{table*}

\subsection{Domain Metrics}
To measure the domain differences, we propose the style, ICV, and IB metrics. 
The assessment of ICV and IB levels are outlined as follows.

\paragraph{\textbf{ICV:}} The distances between different class categories are calculated to measure the ICV. Formally, given a target dataset $\mathcal{D}_{T}$ with categories $\{c_1, c_2, ..., c_N\}$, where $c$ means the name of category e.g., ``cat" and  $N$ denotes the number of class categories, we first extract the text feature $f_{i}$ for each $c_{i}$ using pre-trained CLIP as extractor. The stacked $f_1, f_2, ..., f_N$ constitute the text feature matrix $\mathcal{F} \in \mathbb{R}^{N \times D}$, where $D$ means the dimension of each text feature. Then, we have:
\begin{equation}
       S = \mathcal{F} \cdot \mathcal{F}^T, \quad
      {ICV} = \frac{1}{N^2 * D} \sum_i \sum_j S_{ij}.
\end{equation}

In this way, we get the score matrixes $S_{1}, ..., S_{M}$ and ICV values $ICV_{1}, ..., ICV_{M}$, where $M$ counts the number of target datasets. 
To group the ICV values into three levels, a low bound $ICV_{low}$ and a high bound $ICV_{high}$ are calculated. 
Concretely, for $ICV_{low}$, we choose the minimum value from each score matrix and calculate their average across different datasets; for $ICV_{high}$, we consider the similarity within the same category. 
Therefore, we average the diagonal values of the score matrixes for getting the $ICV_{high}$.  
Formally, we have: 
\begin{align}
     ICV_{low}  = \frac{\sum_k  min\left(S_{k(i,j)}\right)}{M*D} , 
     ICV_{high} = \frac{\sum_k avg \left( diag \left({S_{k}}\right) \right)}{M*D}.
\end{align}

The $diag()$ denotes the diagonal of a matrix, $min()$ means the minimum operation, and $avg()$ means the average operation.  
In this way, we get the $ICV_{low} = 0.112$ and $ICV_{high}=0.190$, and we are able to place the ICV values into their corresponding intervals. 
Results are summarized in Tab.~\ref{tab:icv}. Note that since DeepFish only contains one class, we do not calculate the ICV value for it. 

\begin{table}[h]
\vspace{-0.2in}
 \caption{ICV values and levels of target datasets.}\label{tab:icv}
 \scriptsize
  \centering
  \vspace{-0.1in}
\scalebox{0.9}{
  \begin{tabular}{ccccccccc}
    \toprule
    &  \textbf{ArTaxOr} & \textbf{Clipart1k} & \textbf{DIOR} & \textbf{DeepFish}   & \textbf{NEU-DET}  & \textbf{UODD}  \\
    \midrule
    \textbf{value} &  0.138 & 0.171 &   0.155 & - & 0.183 & 0.132  \\
    \textbf{level} & small & large & medium & - & large & small   \\
    \bottomrule
  \end{tabular}
  }
\vspace{-0.35in}
\end{table}

\begin{figure}[h]
\vspace{-0.25in}
\centering	{\includegraphics[width=.6\linewidth]{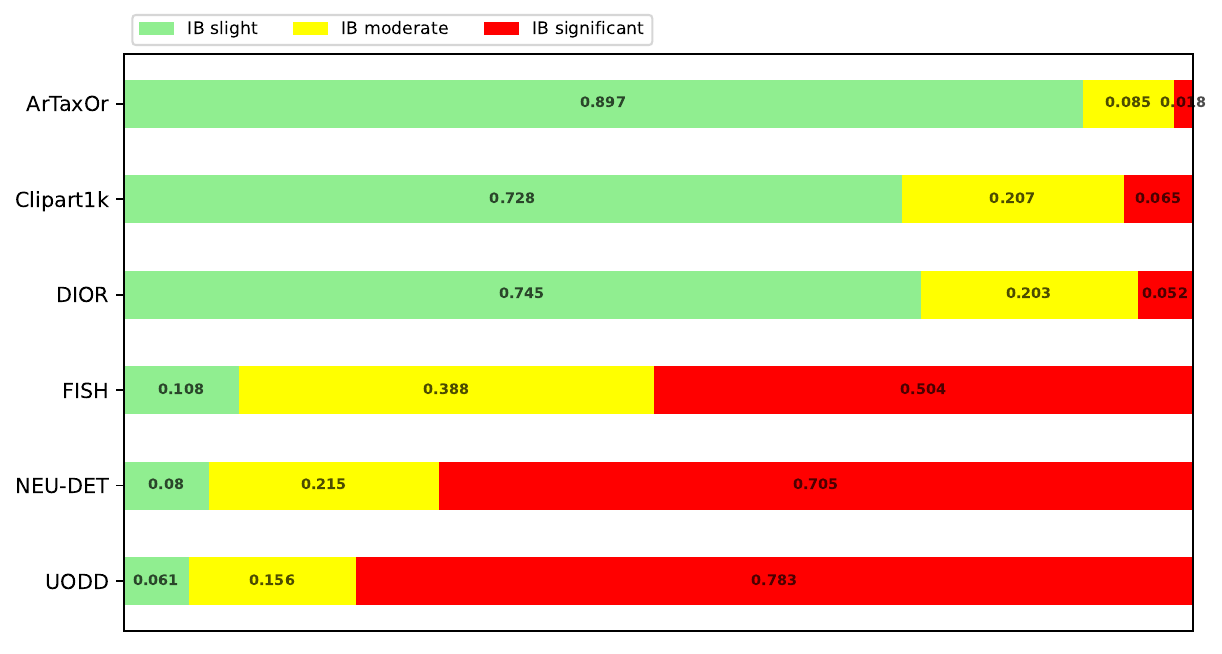}}
\vspace{-0.1in}
\caption{The results of user study on IB levels. Percentages are shown.}\label{fig:supp-ib} 
\vspace{-0.3in}
\end{figure}

\paragraph{\textbf{IB:}} 
While the IB could be assessed by comparing RGB histograms~\cite{fan2021concealed}, 
we opted for a user study to better capture how our human visual system perceives target datasets. We designed a survey questionnaire to estimate the difficulty of detecting target objects from their background, indicating the level of IB. For each target dataset, we randomly selected 10 images, creating a total of 60 images. We then had 100 users classify the IB level into three categories: slight (easy to detect), moderate (somewhat challenging), and significant (hard to detect). The statistical results are summarized in Fig.~\ref{fig:supp-ib}.

Since IB reflects the difficulty of distinguishing foreground and background, we assign the weights of slight $w_{sli}$, moderate $w_{mod}$, and significant $w_{sig}$ as 0, 2, and 6. By calculating the weighted percentages, we got the IB as, 
\begin{align}
     IB = w_{sli}*p_{sli} + w_{mod}*p_{mod} + w_{sig}*p_{sig}.
\end{align}

With IB scores defined, we grouped them into three intervals: [0, 2], [2, 4], and [4, 6].  The final IB values and levels are given in Tab.~\ref{tab:ib}.

\begin{table}[h!]
 \vspace{-0.2in}
 \caption{IB values and levels of target datasets.}\label{tab:ib} 
  \vspace{-0.1in}
 \scriptsize
  \centering
  \begin{tabular}{ccccccccc}
    \toprule
    & \textbf{ArTaxOr} & \textbf{Clipart1k} & \textbf{DIOR} & \textbf{DeepFish}   & \textbf{NEU-DET}  & \textbf{UODD}   \\
    \midrule
    \textbf{value} & 0.278 & 0.804 & 0.718 & 3.800 & 4.660 & 5.010 \\
    \textbf{level} & slight & slight & slight & moderate & significant & significant \\
    \bottomrule
  \end{tabular}
  \vspace{-0.3in}
\end{table}

\end{document}